# Multi-expert learning of adaptive legged locomotion

Chuanyu Yang*[1], Kai Yuan*[1], Qiuguo Zhu[2], Wanming Yu[1], Zhibin Li[1][†]

*These authors contributed equally to this work.

[1] School of Informatics, University of Edinburgh, UK.

[2] Institute of Cyber-Systems and Control, Zhejiang University, China.

[†]*Corresponding author: zhibin.li@ed.ac.uk*

Achieving versatile robot locomotion requires motor skills which can adapt to previously unseen situations. We propose a Multi-Expert Learning Architecture (MELA) that learns to generate adaptive skills from a group of representative expert skills. During training, MELA is first initialised by a distinct set of pre-trained experts, each in a separate deep neural network (DNN). Then by learning the combination of these DNNs using a Gating Neural Network (GNN), MELA can acquire more specialised experts and transitional skills across various locomotion modes. During runtime, MELA constantly blends multiple DNNs and dynamically synthesises a new DNN to produce adaptive behaviours in response to changing situations. This approach leverages the advantages of trained expert skills and the fast online synthesis of adaptive policies to generate responsive motor skills during the changing tasks. Using a unified MELA framework, we demonstrated successful multi-skill locomotion on a real quadruped robot that performed coherent trotting, steering, and fall recovery autonomously, and showed the merit of multi-expert learning generating behaviours which can adapt to unseen scenarios.

## INTRODUCTION

Adaptive motor skills enable living organisms to accomplish sophisticated motor tasks and offer them better chances to survive in nature. Among vast sensorimotor skills, locomotion is essential for most animals to move in the environment. Therefore, to understand and create adaptive locomotion behaviours is a long-standing scientific theme for biologists and roboticists. From a neurological perspective, it is worth understanding how the sensorimotor control system in animals processes various sensory information and produces adaptive reactions in unforeseeable situations (*1–4*). From a robotics perspective, it is interesting to take a bio-inspired approach and transfer biological principles, such as primitive neural circuits, to produce robot behaviours similar to that of animals (*1*). Since the underlying mechanisms of the motor cortex cannot yet be fully replicated (*4*, *5*), we take the latter approach by drawing inspiration from biological motor control to develop learning algorithms which can achieve skill adaptation for robot locomotion (*6*).

In this study, we investigate how an artificial agent can learn to generate multiple motor skills from a set of existing skills, particularly for critical tasks that require immediate responses. As an example of learning a complex physical task, playing soccer consists of several sub-skills, such as dribbling, passing, and shooting. During training, players first practise the most important sub-skills separately. Once mastered, all different sub-skills are used in a flexible combination to improve all these techniques. Our research in multi-skill learning has studied skill-adaptive capabilities such as these, and we draw inspiration from animal motor control when designing the learning and control architecture. Using a quadruped robot as a testbed, we aim to produce adaptive robot behaviours to succeed in unexpected situations in a responsive manner.

## Background

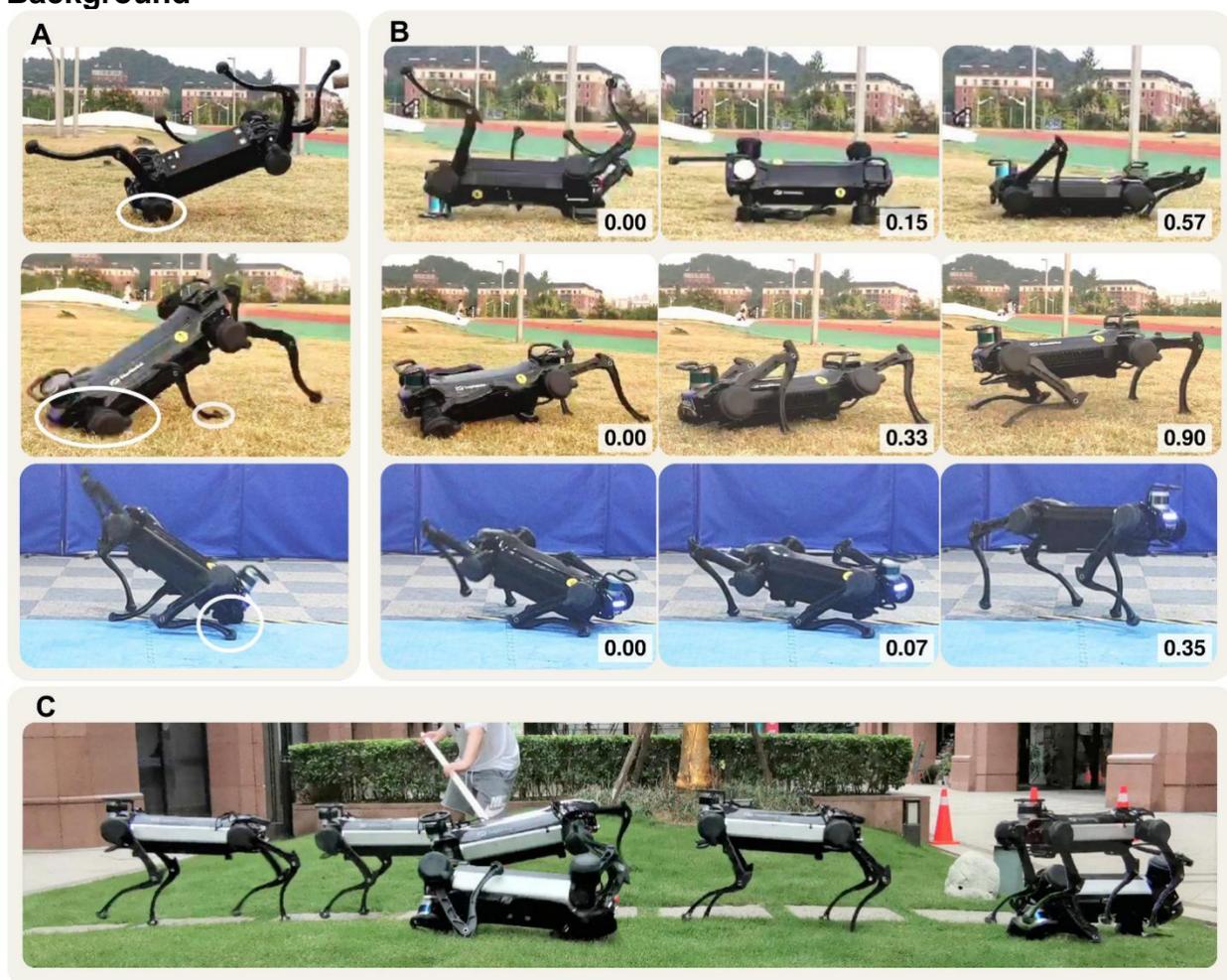

**Fig. 1. Challenging locomotion scenarios and agile manoeuvres of a quadruped robot.** (**A**) Three challenging scenarios of the *Jueying* robot during various tests: unexpected body contacts with the environment and unpredictable robot states. The white circled regions highlight unusual contact that can occur at any part of the body. (**B**) Different adaptive behaviours from our proposed learning framework that generated dynamic motions and complex coordination of legs for immediate recovery from failures. (**C**) Resilient locomotion using adaptive skills in presence of unexpected disturbances. (Time in snapshots is measured in seconds).

The DARPA Robotics Challenge (DRC) from 2012 to 2015 fostered the development of semi-autonomous robots for dangerous missions such as disaster response in unstructured environments (*7*). Most DRC robots had different forms of legged design for the dexterity to traverse irregular surfaces. Despite the tremendous engineering efforts, no robot could recover from falls autonomously (*8*). To date, most legged robots still lack such an autonomous ability to generate adaptive actions to deal with unexpected situations.

Due to the uncertainties in unforeseen situations, locomotion failures are likely to happen. We illustrate these challenges in robot locomotion using field tests (Fig. 1A) and the adaptive behaviours that are robust to uncertainties (Fig. 1B-C). Typically, falling occurs within a second of the robot losing balance and the time window for fall prevention is around 0.2 - 0.5 seconds. Therefore, it is critical to immediately coordinate different locomotion modes to mitigate perturbations, and prevent or recover from failures. In comparison to robot systems, biological systems, e.g., cats, dogs, and humans, exhibit higher versatility (*9*), and the key to the performance gap is the difference in motion intelligence which allows biological systems to handle changing and complex situations (*9–11*).

Our paper studies a machine learning approach that learns reactive locomotion skills and generates adaptive behaviours by reusing and recombining trained skills. Here, we investigate the motor skills in the form of feedback control policies to address the reactive adaptation to multimodalities during robot locomotion, leading to increased robustness against failures.

**Related work**
When a robot interacts with its environment, it can be difficult to determine which part of the robot is in contact; this is challenging for classical control solutions, as careful modelling of contacts is often needed. The main approach in the legged locomotion community uses model-based mathematical optimisation to solve these multi-contact problems, such as Model-Predictive Control (MPC), whole-body Quadratic Programming (QP), and Trajectory Optimisation (TO). To achieve fast online computation, MPC utilises simplified models and short predictive horizons to plan task-space motions for walking (*12*), running (*13*) and pacing (*14*). QP methods are used for whole-body control to map task-space motions (e.g., those from the MPC) to joint-space actions while considering the whole robot model and physical constraints (*15, 16*). A unified but computationally expensive technique is to optimise all models and constraints together (whole robot model, contact model, environment constraints), i.e., through nonlinear MPC (NMPC) (*17*) and whole-body optimisation (*18–20*).

In the optimisation scheme, all physical contacts between the robot and the environment need to be defined as constraints in the formulation. The contact sequence, such as the contact location, timing, and duration, needs to be specified either by manual design or by an additional planner (*21, 22*). Furthermore, the explicit properties of the robot and the environment need to be modelled (*23*), but are expensive to compute (*24*) and thus difficult to run in real-time in complex settings even with exhaustive computing (*25*). This fundamental principle suffers from the curse of dimensionality, and therefore limits the scalability to real-time solutions in more complex and challenging problems (*26*).

To this end, Deep Reinforcement Learning (DRL) becomes attractive for acquiring task-level skills: through rewarding intended outcomes and penalising undesired ones, an artificial agent can learn desirable behaviours (*21, 22*). Using DRL offers several advantages: the training process can be realised by using physics engines to perform a large number of iterations in simulations without risks of hardware damage; the agent can explore freely and learn effective policies that are difficult for humans to manually design; and the computation of readily trained neural networks can be real-time. For legged locomotion, many DRL results have been achieved in simulation (*27, 28*) and, in recent studies, on hardware (*29–31*), e.g., demonstration of learning-based control on a real robot using separate policies for fall recovery and walking (*32*). However, similar to other DRL approaches, the learning policies in (*32*) were specialised in separate tasks instead of being a unified policy across different tasks. This is a common feature due to the learning structure that only trains a single DRL agent for solving one specific task, which results in a narrowly skilled policy.

Hierarchical Reinforcement Learning (HRL) solves complex tasks at different levels of temporal abstraction using multiple experts' existing knowledge (*33*): experts are trained to encode low-level motor primitives, while a high-level policy selects the appropriate expert (*34, 35*). However, generating new skills cannot be achieved in the standard HRL framework since only one expert is selected at a time. This problem can be addressed by learning to continuously blend high-dimensional variables of all experts (*36*). One related approach is the Mixture of Experts (MoE) that synthesises the outputs of individual experts specialised on sub-problems using a gating function (*37*), which has been used in robotics (*38*), computer vision (*39*), and computer graphics (*40*). However, compared to blending the experts' high-dimensional variables, MoE has known limitations in scaling to high degree-of-freedom systems (*40*), and exhibits expert imbalance problems, i.e., favouring certain experts while degrading others (*41*).

**Our approach**
We draw inspiration from biological motor control to design our control and learning framework. Biological studies suggest that motor behaviour is controlled by the Central Nervous System (CNS) that resets the reference position of body segments, and the difference between the reference and the actual position excites the muscular activities for generating appropriate forces (*42*). This precludes the need to compute the inverse dynamics, simplifies control and minimises the computation (*43*). Since the spring-damper property provided by the impedance control resembles the elasticity of biological muscles, we applied the Equilibrium-Point (EP) control hypothesis to generate joint torques by offsetting the equilibrium point.

Inspired by the biomechanical control of muscular systems and the EP hypothesis, we distribute the robot control in two layers: (i) at the bottom layer, we use torque control to configure the joint impedance for the robot; and (ii) at the top layer, we designate deep neural networks (DNNs) to produce set-points for all joints to modulate posture and joint torques, establishing force interactions with the environment (see Materials and Methods). By doing so, we can focus on developing the learning algorithms at the top layer to achieve motor intelligence.

**Contributions**

This work aims to demonstrate how hierarchical motor control using deep reinforcement learning can achieve a breadth of adaptive behaviours for contact-rich locomotion. We propose a Multi-Expert Learning Architecture (MELA) which contains multiple expert neural networks, each with a unique motor skill, and a gating neural network that fuses expert networks dynamically into a more versatile and adaptive neural network.

Compared to the approach of using kinematic primitives (*44*, *45*), the proposed MELA policy indirectly modulates the joint torques by changing the reference joint angles, where the resulting motions are the natural outcomes of the dynamic interactions with the environment. In contrast to other hierarchical learning approaches that select one policy representing one skill at a time (*34*, *35*), MELA continuously combines network parameters of all experts seamlessly, and is therefore more responsive than other methods since there is no wait time due to disjointed switching of experts. Additionally, since MELA synthesises the experts in a high-dimensional feature space, i.e., weighted average of the network parameters (weights and biases of the neural networks), it does not suffer from the same expert imbalance problems as MoE (*41*). Similar multi-expert structures that blend experts in high-dimensional feature space were studied in computer graphics for kinematic animation (*41*, *46*, *47*), but have not yet been developed as feedback policies for the control of dynamical systems such as robots.

Our presented work contributes to a learning framework that (i) generates multiple distinctive skills effectively, (ii) diversifies expert skills using co-training, and (iii) synthesises multi-skill policies with adaptive behaviours in unseen situations. The adaptive behaviours are verified by proof-of-concept experiments on a real robot and various extreme test scenarios in a physics simulation. By synthesising multiple expert skills, the collective expertise of MELA is more versatile than each single expert, thanks to the dynamic structure of MELA, which integrates all experts based on the online state feedback. Such multi-expert learning allows each expert to specialise in unique locomotive skills, i.e., some prioritise postural control and failure recovery while others acquire strategies for maximising task performance. As a result, MELA can perform a broad range of adaptive motor skills in a holistic manner, and is more versatile because of the diversification among experts.

The proposed framework is effective in achieving reactive and adaptive motor behaviours to changing situations consisting of unforeseen scenarios, unexpected disturbances, and different locomotion modes, which have not been addressed well by other unified frameworks in the literature. We will show the advantages of our proposed work which can produce a variety of strategies to ensure task success. For example, the robot can produce quick responses to recover balance in different unexpected postures and the ability to stabilise dynamic transitions. This is an indicative level of machine intelligence – the ability to autonomously produce locomotive skills that would require significant intelligence to design if they needed to be programmed by humans.

## RESULTS
### Multi-expert learning framework

We first define key terminology to help explain the concepts in this article: motor skill, expert, and locomotion mode. *Motor skill (or "skill" for short)*: a feedback policy that generates coordinated actions to complete a specific type of task; this serves as a building block for constructing more complex manoeuvres. *Expert*: a deep neural network with specialised motor skills. *Locomotion mode*: a pattern of coordinated limb movement, such as standing, turning on the spot, trotting forwards and backwards, steering left and right, and fall recovery.

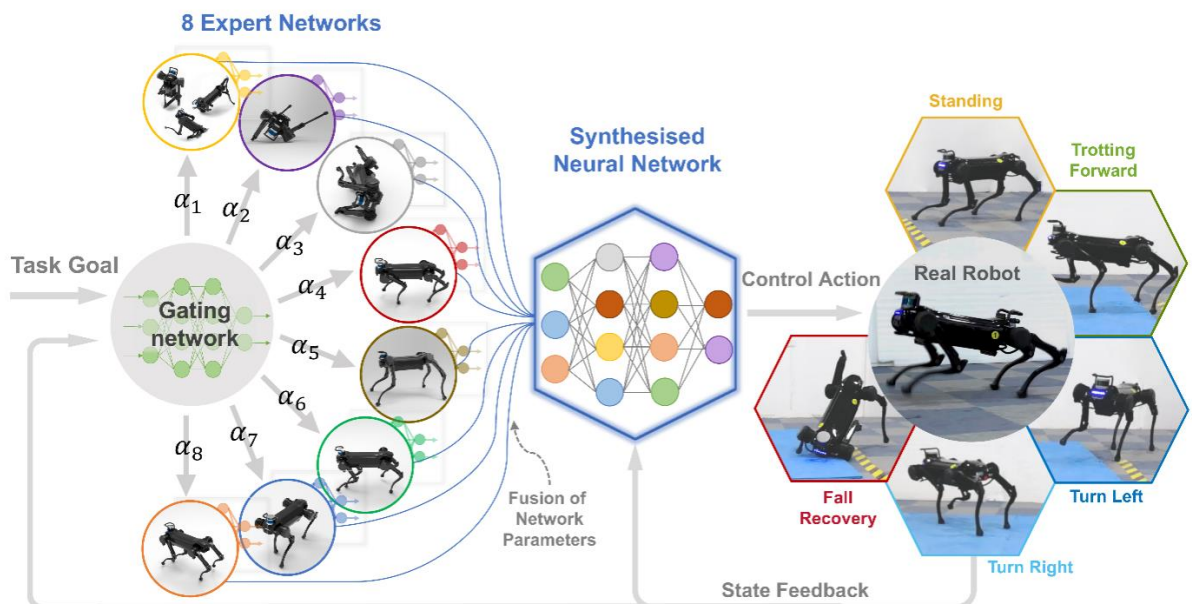

**Fig. 2. Multi-Expert Learning Architecture (MELA): a hierarchical deep reinforcement learning framework that generates adaptive behaviour by combining multiple deep neural networks (DNNs) together to produce versatile locomotive skills.** The Gating Neural Network (GNN) generates variable weights (α) to fuse the parameters of all eight expert networks (each expert is illustrated by its primary motor skill), such that newly synthesised motor skills are adapted to different locomotion modes by blending useful learned behaviours collectively from the consortium of experts.

To complete tasks in unseen scenarios, an artificial agent needs the ability to adapt relevant skills during runtime, which is why we propose the use of MELA: a hierarchical reinforcement learning structure which consists of a collection of Deep Neural Networks (DNNs) and a Gating Neural Network (GNN). As shown in Fig. 2, the GNN continuously fuses expert DNNs into a single synthesised neural network at every time step by computing the weighted average of all experts' network parameters. The synthesised policy fully encodes the motor skills of the experts in a high-dimensional feature space (see Materials and Methods). Through repurposing and combining existing skills, MELA acquires a wide array of adaptive behaviours and achieves versatile locomotion in unseen scenarios.

The multi-expert policy of MELA is trained in a two-step process. In the first stage, we train a set of policies, where each one accomplishes a distinct task. Specifically, we use the trained experts, such as fall recovery and trotting experts, to initialise all experts by two sub-groups. In the second stage, we use a GNN as the merging mechanism to fuse all network parameters and train all experts together, so that their collective specialisations can be fully utilised by the gating network. Meanwhile, the gating network is trained to learn the continuous and variable activation of different experts to generate optimal policies at each control loop (see Materials and Methods). The MELA policy was trained in the physics simulation and evaluated on a real robot system.

For constructing MELA, the number of experts is determined by the relation between the desired locomotion tasks and the required skills. A particular motor skill corresponds to a distinct control strategy (e.g., to roll the body by pushing the ground), and thus one locomotion task would require a set of different skills. In our study, we focus on five locomotion tasks (Fig. 2): fall recovery, standing, turning left, turning right, and trotting. All these movements need a variety of motor skills for interacting with the environment.

According to the configurations that the robot would encounter during these five tasks, a basic number of experts can be determined. There are 7 *distinct situations* each requiring at least one motor skill, namely: (i) fall recovery from a supine pose (lying on the back), (ii) fall recovery from lateral decubitus poses (lying on the left/right side), (iii) balance control during stance, (iv) body postural control, (v) trotting forward, (vi) left steering, and (vii) right steering. Hence, a minimum of 7 DNNs can be used to represent these skills. We also introduced a redundant expert that can represent nonlinear features and additional skills which are difficult to anticipate. As a result, we constructed the MELA using 8 experts in total. Further comparison found that using more than 8 experts had no further performance but more training time (fig. S1). It shall be noted that the 7 situations are only used as a guide to determine the number of experts at the initial design process, and does not need to match the order of numbering of MELA's learned motor skills in the latter section.

**Learning individual motor skills – Fall recovery and trotting**
In challenging physical tasks, it is difficult to directly train control policies where a variety of skills are needed, such as recover falling poses and resume walking gaits. Prior studies show that pre-learning skills allow the experts to learn task representations, otherwise the policies were not able to learn to solve more difficult composite tasks (*36*, *40*). Like training for sports, it is essential to practise individual skills that are distinctly different from each other. Similarly, MELA has a two-step training procedure for experts, in which two distinct, separate modes are specialised first: fall recovery and trotting. In the following, we will show the experimental results of fall recovery and trotting using individually trained neural networks, and then present the multimodal locomotion experiments using MELA.

For quadrupeds, a canonically stable configuration is a standing posture with four feet forming a support polygon close to the body's length and width. For fall recovery, the DRL agent is rewarded for feedback policies that restore such stable postures from various failure states. We applied random initial configurations to explore diverse robot states and facilitated the agent's ability to generalise policies for various fall poses (see Supplementary Materials).

We evaluated the robustness of recovery policies and categorised the learned reactive behaviours into four strategies (Fig. 3A-C): (i) natural rolling exploiting semi-passive movements, (ii) active righting and flipping, (iii) standing up from prone positions, and (iv) stepping. *Natural rolling* describes the behaviour where the robot exploits its natural dynamics and gravity to roll over. This is activated when the robot is in a prone and/or lateral decubitus position, as shown in Fig. 3A. *Active righting* is the strategy where the robot pushes itself using the leg and elbow, creating momentum to actively flip itself to a prone position, as shown in Fig. 3B. *Stepping* emerges when necessary during standing to regain balance, involving coordination and switching of support legs. An example of stepping is shown in Fig. 3C, and such multi-contact switching was all generated naturally using the learned policy based on the online state feedback.

From all fall recovery experiments (Fig. 3A-C and movie S1-2), we can see the responsive and versatile reactions, including the emerged stepping behaviour. Compared to baseline fall recovery which is manually engineered with a fixed pattern (fig. S2), our learning policy is able to recover from various fall scenarios because it can respond to dynamic changes using online feedback, while the handcraft controller only addresses a narrow range of situations.

The *Jueying* robot can robustly trot under three ground conditions with different stiffness, friction, and obstacles (Fig. 3D-F and movie S3). In Fig. 3D, the concrete ground was covered by thin carpets with high friction, while in Fig. 3E, 2*cm* thick foam mats were laid, creating a softer and more slippery surface. In Fig. 3F, 5*cm* thick bricks were scattered as small obstacles. The learned motor skills were robust under different ground conditions, and the *Jueying* robot was able to continue trotting steadily in all three scenarios.

All these trained policies have exhibited behaviours of compliant interaction to handle physical interactions and impacts. The joint impedance mode offers the ability to indirectly regulate joint torques using the deviation between the desired joint position $q^d$ and actual joint position $q^m$ via the principle similar to the series elastic actuators, i.e., $\tau = K_p(q^d - q^m)$ (*48*). The expert has indirectly learned active compliance control by regulating the references based on feedback of the current joint positions to minimise joint torques.

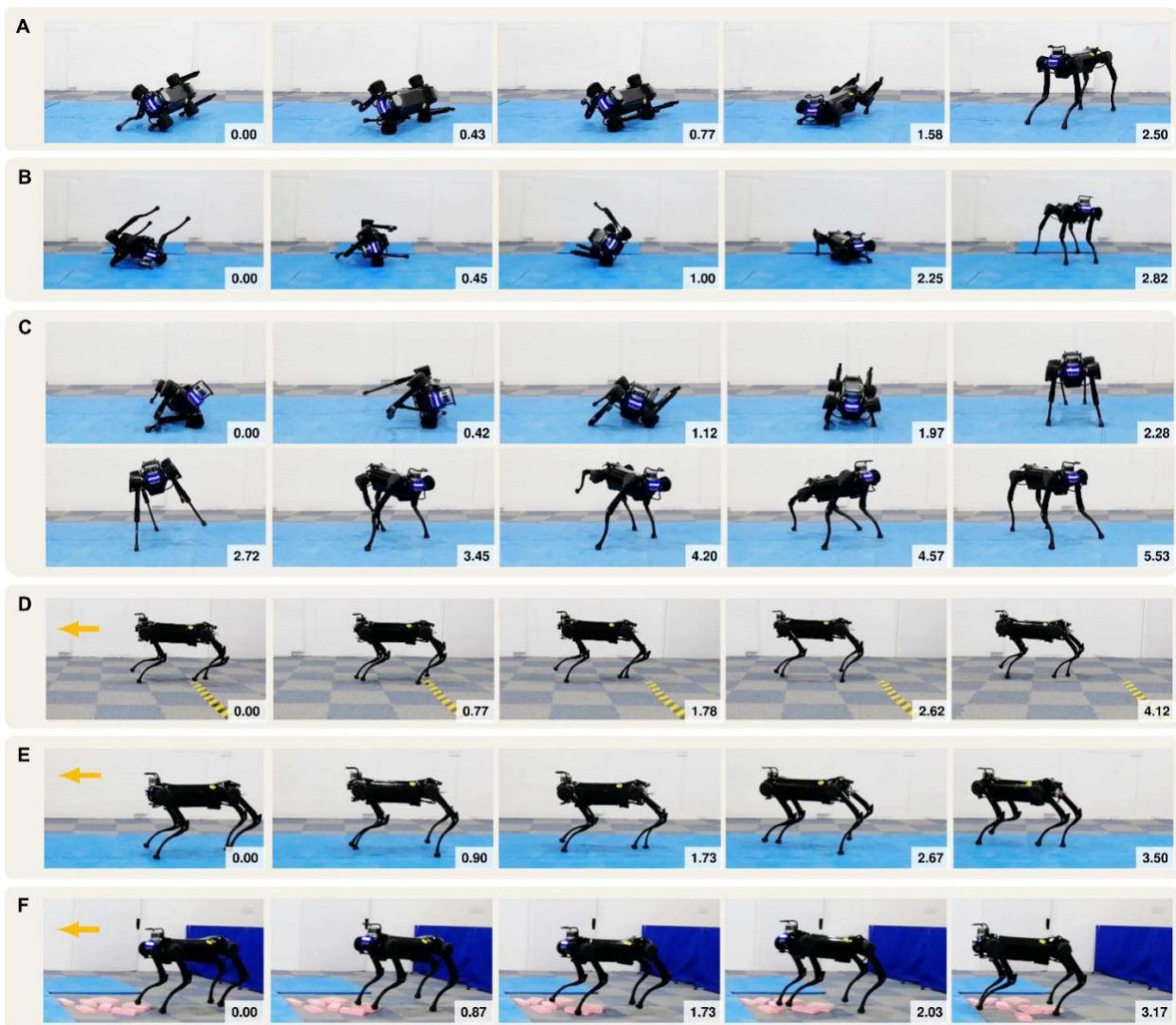

**Fig. 3. Individual motor skills for the fall recovery and trotting respectively.** (**A**) A configuration between prone and lateral decubitus positions where legs were stuck underneath the body: the robot first pushed the ground to lift up the body for ground clearance, and then retrieved legs to a prone posture for standing up. (**B**) The robot actively used elbow-push to generate a large momentum to self-right to a prone position. (**C**) A stepping behaviour was learned and performed naturally to keep balance. (**D**) Stable trotting on a hard floor. (**E**) Stable trotting on soft slippery foam mats. (**F**) Stable trotting over scattered obstacles, showing the compliant interaction and robustness learned by the trotting expert. (Time in snapshots is measured in seconds).

## Multi-expert skills for multimodal locomotion
### *Analysis of learned MELA policy*

After the first stage of the two-stage training process, the network parameters of fall recovery and trotting policies are transferred into the expert networks in MELA. In the second stage, all experts are co-trained with the gating network, and MELA repurposes the initial experts to learn adaptive behaviours necessary for multimodal locomotion, while the gating network learns how to blend the acquired skills to respond to the changing tasks. As a result, MELA is able to achieve non-cyclic and asymmetric motions (fall recovery), rhythmic movements (trotting), goal-oriented tasks (target-following), as well as all dynamical transitions between different modes. These key adaptive behaviours of the MELA policy were tested on a real robot, which demonstrated the capabilities of achieving a diversity of locomotion tasks, adapting to external environmental changes responsively, whilst also following user commands.

To analyse the inner workings of MELA, we performed systematic tests of the trained MELA networks in the physics simulation, so as to fully cover the wide range of sensory inputs. Without hardware risks, we operated the robot in extremely dynamic motions to activate different experts, allowing analysis of all distinct motor skills. This provides data of variable weights that reflects the activation level of all experts over each motor skill. Figure 4A shows the correlations and patterns between the activation of experts and the motor skills, and reveals that each motor skill has a dominant expert, suggesting the primary specialisation of each MELA expert.

As shown in Fig. 4A, *8 fundamental motor skills* are acquired: (i) back righting, e.g., push elbow to roll over from supine positions (lying on the back); (ii) lateral rolling, e.g., retrieve legs and roll the body to a prone position (lying on the abdomen); (iii) postural control, e.g., maintain a nominal body posture; (iv) standing balancing, e.g., maintain stable stance and take steps when necessary; (v) turning left; (vi) turning right; (vii) trotting at small steps; and (viii) trotting at larger steps. These *8 fundamental motor skills* are the building blocks for MELA to compose variable skills and produce adaptive behaviours.

The skill specialisation and distribution among experts emerge naturally through the MELA co-training. Therefore, the order of the experts does not need to follow the numeration of the specialised motor skills. The primary motor skill specialisation of experts 1 to 8 are: turning right (skill vi), standing balance (skill iv), large-step trotting (skill viii), turning left (skill v), body posture control (skill iii), back righting (skill i), small-step trotting (skill vii) and lateral rolling (skill ii). As the result of introduced redundancy, expert 7 was exploited and trained as a complementary role in conjunction with expert 3 for trotting forward, i.e., expert 7 and 3 were specialised in trotting at small and large steps, respectively. An alternative visualisation can be found in fig. S3 where experts are sorted by activation patterns across different motor skills.

### *Analysis of skill adaptation and transfer*
Apart from the skill specialisation (Fig. 4A), we studied how skills are adapted and transferred in the MELA networks by using t-distributed Stochastic Neighbour Embedding (t-SNE) to analyse coactions of the gating network and the expert networks, respectively. The t-SNE algorithm is a dimensionality reduction technique to embed and visualise high-dimensional data in a low-dimensional space. It first computes a conditional probability distribution, representing the similarities of samples in the original high-dimensional space based on a distance metric, and then projects samples to a low-dimensional space in a probabilistic manner. Therefore, similar output actions from the networks will appear with high probability in the same neighbourhood as clustered points (Fig. 4B-E), and vice versa.

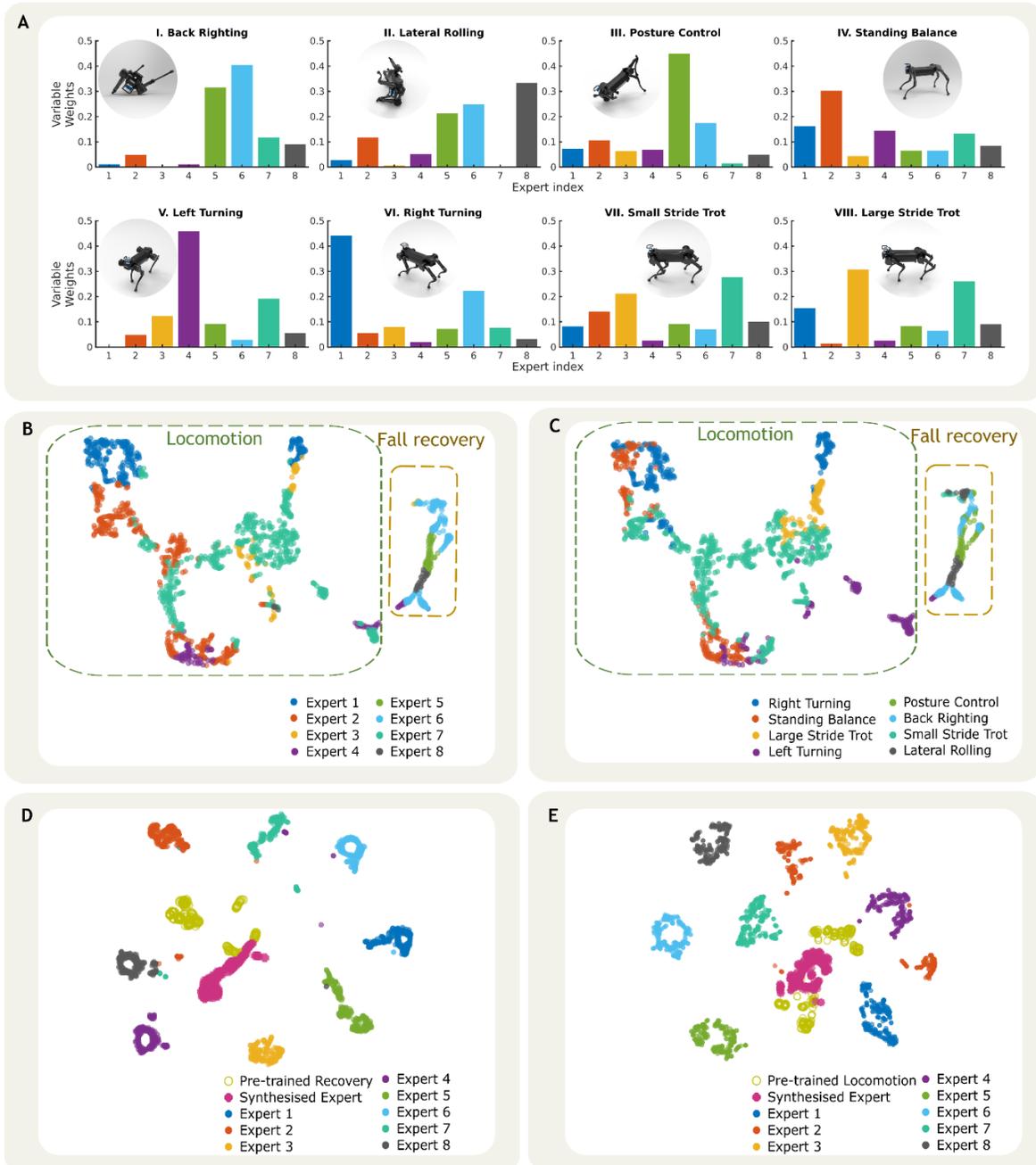

**Fig. 4. Analysis of the specialisation of experts across different motor skills, and the patterns of the gating network and the expert networks using the t-distributed Stochastic Neighbour Embedding (t-SNE).** The analysis was based on the simulation tests using the policy from a single training run, which was representative because the training process can consistently reproduce policies with very similar characteristics of skill specialisation. (**A**) Specialised activation of eight experts across different motor skills, where the distinct activation patterns indicate a unique specialisation (table S1). (**B-C**) The 2D projection of the gating network's activation pattern by t-SNE, where the neighbourhoods and clusters of the samples are visualised. Samples representing similar activation appear in close proximity, whereas the different ones are distant from each other. (**B**) The output samples of the gating network, which are classified by the index of the dominant expert that has the highest activation (Fig. 4A). (**C**) The output samples of the gating network, which are classified by the physical states during a distinct locomotion mode, e.g., trotting,

balancing, turning left/right. (**D-E**) The 2D projection of the actions from the pre-trained, co-trained, and synthesised expert policies using t-SNE analysis: (**D**) and (**E**) are the target actions classified during fall recovery and trotting tasks, respectively.

In Fig. 4B-C, the t-SNE analysis on the outputs of the gating network (the variable weights for all experts) reveals the relationship between experts and the resulting motor skills, and how the gating network synthesises experts after the MELA learning process. There are two maps of clusters in both Fig. 4B and Fig. 4C, which are grouped by dashed lines corresponding to locomotion (green) and fall recovery (bronze), suggesting that the gating network perceives these two as different modes. The t-SNE analysis is labelled according to the experts (Fig. 4B) and motor skills (Fig. 4C), respectively, where the clustered samples have matching distributions mostly between these two maps, indicating each expert's primary motor skill is in agreement with the activation patterns shown in Fig. 4A.

We also compared actions generated by all experts using t-SNE and revealed how multiple skills evolved and diversified after co-training. As shown in Fig. 4D-E, the actions generated from eight experts are distant from each other, meaning that experts have been specialised towards more unique skills. The limited intersection between the clusters of the trained experts in stage 2 and the initial experts from stage 1 also implies that: the trained experts have diversified from the original skills and further acquired more profound and newly emerged skills during MELA's co-training stage. The data in Fig. 4D-E show interesting results: the cluster of actions from the synthesised network intersects with those from the pre-trained expert policies, meaning newly emerged behaviours of fall recovery and locomotion share some similarities with the original ones; the dynamically synthesised expert partially preserves the original skills, which are reconstructed by fusing eight distinctive experts.

### *Multi-skill locomotion*
To validate the performance of the MELA policy, we designed experiments that were safe to execute on the real robot with an increasing number of locomotion modes: (i) single-mode fall recovery (Fig. 5A); (ii) double-mode left-right steering on the spot (Fig. 5B); (iii) triple-mode of simultaneous left-right steering and trotting (Fig. 5C); and (iv) target-following locomotion involving all modes, i.e., standing, left-right steering, trotting and fall recovery (movie S4-5).

In our study, adaptive behaviours refer to the online synthesised skills that adapt reactively to previously untrained situations. We summarise the adaptive behaviours achieved by MELA in two categories: (i) *Emerged skills* that are newly acquired during training in stage 2 of MELA, i.e., skills for steering and turning (Fig. 5 and fig. S4A) and variable-speed trotting (fig. S5); (ii) *Transitional skills* that coordinate dynamical transitions smoothly between different locomotion modes, e.g., transition from various failure poses to trotting (Fig. 5 and fig. S4B-E). Five representative cases of the adaptive behaviours from the MELA policy can also be found in fig. S4.

Figure 5A shows successful fall recovery performed by the MELA policy, and the similarity with those in Fig. 3A-C indicates that the MELA policy has reused some pre-trained skills. Figure 5B shows that the MELA policy was able to infer the heading direction from the target location, and learned how to perform swift turning to track the changing target. During the left and right steering experiments (Fig. 5B), the average turning velocities were 1.6 *rad/s* (92.0 *deg/s*) and -1.1 *rad/s* (-61.7 *deg/s*) with peak values at 2.7 *rad/s* (156.8 *deg/s*) and -2.7 *rad/s* (-156.9 *deg/s*), respectively (fig. S6). Although the experts initialised in MELA were only for trotting and fall recovery, MELA was able to reshape the existing experts for the steering tasks as one of the newly emerged skills.

Figure 5C and 5D show target-following tasks requiring simultaneous trotting and steering on the real robot. The task was to chase a virtual target given by a user command, i.e., a variable position vector with respect to the robot. In Fig. 5C, a small target position ahead of the robot was provided, e.g., 0.28 m in the heading direction (fig. S7A), and the robot performed left/right turning while trotting forward. In the experiment shown in Fig. 5D, a farther target position of 0.48 m was commanded (fig. S7B), and the robot was chasing a distant target and trotting at larger steps. Torque saturation of motors occurred more often on the lab floor (fig. S8), and the robot had three successive tripping and recovery incidents. Key snapshots around falling and reactive responses are in Fig. 5D. Once the states of gait failures were sensed, MELA was able to produce immediate reactions to restore balance within one second (fig. S9), and the robot recovered from tripping and continued trotting without human intervention. Figure 5E shows an outdoor experiment where the robot first recovered from a falling posture and was later knocked down during a long walk on the grass. MELA was able to recover the robot from fall and resume trotting successfully (fig. S10). From all experiments, the synthesised MELA expert demonstrated flexible motor skills, coordinated movements and smooth transitions, showing how crucial it is to have such reactive responses and feedback control for robust and autonomous locomotion.

As shown in Fig. 5, MELA enabled the robot to complete all validation successfully, and demonstrated dynamic fall-resilient locomotion (see more in fig. S11). The gating network in MELA has learned how to generate variable weights for all experts in response to the state feedback and to provide smooth transitions across all modes, and see data analysis of Fig. 5D-E cases in fig. S12 and fig. S13. Meanwhile, all the trained experts were activated coherently to collaborate with each other under the regulation of the gating network in order to synthesise an optimal skill suited for the situation. Additional analysis of the gating pattern and the relationship between experts is presented in the Supplementary Materials.

To further evaluate the performance, the MELA policy was validated by additional test scenarios in simulation that were not encountered during training, including gravel, inclined surfaces, a moving slope, rough terrain (fig. S14) as well as robustness tests with variations of masses and motor failures (fig. S15). During successful locomotion in these unseen scenarios, the MELA policy performed versatile adaptations to unexpected situations (movie S6 and S7). We note that the MELA framework has learned to deal with various transitions between different locomotion modes (fig. S16-20), and also the synthesised policy is different from the eight basic motor skills which indicates a nonlinear interpolated behaviour among expert skills (fig. S21). All these experiments and simulations validate MELA's capability of producing flexible behaviours in a variety of unforeseeable scenarios.

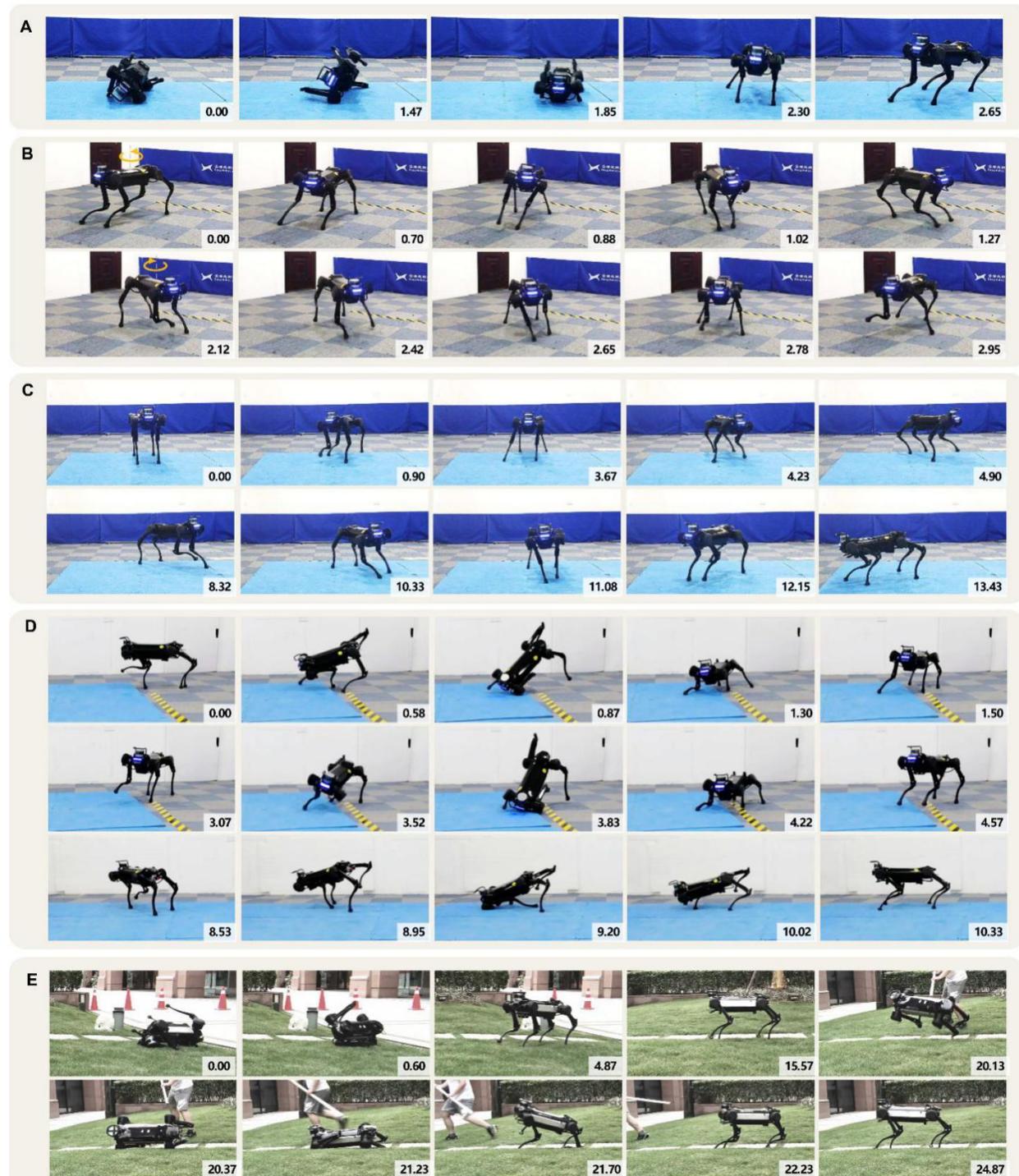

**Fig. 5. Dynamically synthesised MELA policy running on a real quadruped robot and demonstrating adaptive and agile locomotion behaviours.** (**A**) Successful fall recovery performed by the MELA expert,

inheriting original skills from the pre-trained experts. (**B**) Newly emerged skills, dynamic steering on the spot, naturally learned through the MELA framework. (**C**) Target-following experiment with simultaneous trotting and steering using online synthesised skills. (**D**) Target-following experiment showing the capability of failure-resilient trotting and critical recovery within one second (averagely 0.5 second for restoring body posture and 0.4 seconds for returning to the trotting mode). (**E**) Adaptive trotting and fall recovery experiment on a sloped grass field, showing the capability of robust traversal in unstructured environments. (Time in snapshots is measured in seconds).

## DISCUSSION

This study aims to achieve versatile robot motor skills in contact-rich multimodal locomotion. In contrast to most solutions which are dedicated to separate narrow-skilled tasks, we approached this challenge with a hierarchical control architecture of multi-expert learning – MELA – which is able to generate adaptive motor skills and achieve a breadth of locomotion expertise. In particular, MELA learns to generate adaptive behaviours from trained expert skills by dynamically fusing a new synthesised neural network, i.e., a feedback policy that reacts quickly to changing situations. This is essential for autonomous robots to respond rapidly in critical conditions and is more useful for mission success in real-world applications.

In comparison with MoE, MELA's approach of fusing network parameters prevents the expert imbalance problem and provides diversity among expert skills. As a result, all experts are required to have the same neural network structure for implementing MELA. The training of MELA is a two-stage process with an initialisation of fall recovery and trotting policies at the first stage, and the multi-expert co-training with the gating network at the second stage. The t-SNE analysis of all expert networks and the gating network suggest that: the consortium of multiple experts have expanded the initial pre-trained expert skills and acquired more distinct and diverse skillsets; the high-level gating network has learned to distinguish each expert and blend the weights of each specialisation according to different conditions; and the composed synthesised MELA expert partially preserves some of the original skills. We implemented the MoE approach using the same pre-trained experts and training procedure, and the MoE policy exhibited degenerated experts in steering skills which led to asymmetric gait and turning behaviour (note S3 and fig. S22).

The experimental and simulation results outline MELA's key contributions in learning a variety of adaptive behaviours from specialised experts, the adaptation to changing environments, and the robustness against uncertainties. The experimental results show that MELA achieved multimodal locomotion with agile adaptation and fast responses to different situations and perturbations, i.e., smooth transitions between standing balancing, trotting, turning, and fall recovery. As a learning-based approach, MELA leverages computational intelligence and shows the advantage of generating adaptive behaviours compared to traditional approaches that purely rely on explicit manual programming.

### Limitations and future work

Though our current MELA scheme is able to generate adaptive policies, it has no visual and haptic perception which are critical for long-term motion planning (*49*), dynamic manoeuvres (*50*), and utilisation of affordance to coordinate whole-body support poses (*51*). To acquire more advanced motion intelligence in unstructured environments, future research needs to integrate visual cues and haptic sensing to develop environment-aware locomotion.

While scaling up the number of locomotion modes, training in physics simulation may impose some limitations. Though all policies were validated by the *Jueying* robot in five locomotion modes, the discrepancy between the simulation and the real world may accumulate and arise as an issue, when the number of tasks increases. Since the scope of this research is to achieve a diversity of reactive skills rather than sim-to-real transfer, we performed training in simulation and avoided potential damage to the real 40kg-robot during the exploration of the learning algorithm. Based on the results of MELA, the future work will be on the learning algorithms that can refine motor skills safely on real hardware for more complex multimodal tasks.

## MATERIALS AND METHODS

In this section, we will first introduce the robot platform and then explain the core design of the MELA framework, including reward terms, state observations, and action space. Particularly, we will present an emulation of the frequency response of actuators and a loss function design for producing smooth and feasible actions. Finally, we elaborate on the MELA framework and the 2-stage MELA training procedure.

### Robot platform

We implemented our learning algorithms on the *Jueying* quadruped robot (*52*) to validate the adaptive behaviours with feasible and safe tests on the real hardware. *Jueying* has 3 degrees of freedom (DoF) per leg (12 DoFs in total) which are actuated by brushless electric motors with low gear ratio (i.e., 7:1) and high-fidelity joint torque control (table S2).

### Deep reinforcement learning framework

The goal of the DRL is to train an artificial agent to infer optimal actions from the current state by learning from past experience. The experience samples are stored as tuples containing the current state $s_t$, the action generated from policy $a_t \sim \pi_t(s_t)$, the reward $r_t$, and the next transition state $s_{t+1}$. These samples are first collected and stored in a Replay Buffer, from which they are drawn later to train the policy. We used the Soft Actor Critic (SAC) algorithm (*53*), and below are the details of reward, state, action, training procedures and loss function.

### *Reward design*

For training individual tasks, such as fall recovery, trotting and target-following, we designed a specific reward function with corresponding weights for reward terms that represent different physical quantities. The full list of reward terms is: (i) base pose, (ii) base height, (iii) base velocity, (iv) joint torque regularisation, (v) joint velocity regularisation, (vi) body ground contact, (vii) foot ground contact, (viiii) yaw velocity, (ix) swing and stance, (x) average foot placement, (xi) reference joint position, (xii) reference foot contact, (xiii) robot's heading to the goal, and (xiv) the goal position. Different tasks require a specific subset of reward components, i.e., fall recovery requires reward (i) to (vii), trotting requires reward (i) to (xii), and multimodal target-following locomotion requires all 14 reward terms. In summary, the first 7 reward terms are common physical quantities across all tasks to ensure stable robot motion, while the other terms are task-specific. The mathematical formulation of all reward terms and the task-specific weights are in table S3 and table S4.

### State observation

We used the following state observations that are essential and minimalistic to train successful policies: (i) base (robot body) orientation, (ii) angular velocity of the robot base, (iii) linear velocity of the robot base, (iv) joint positions, (v) phase vector, and (vi) goal position (*54*). The body orientation is represented as a normalised gravity vector projected in the robot local frame using the measurements from the Inertial Measurement Unit (IMU). The angular velocities of the body and all the joint positions were measured by the IMU and joint encoders, respectively. The linear velocity was obtained from the state estimation by fusing the leg kinematics and the accelerations from IMU (as a strap-down inertial navigation system) and then transformed to the heading coordinate, so the resulting velocity is agnostic to the heading direction. The 2D phase vector was designed to clock along the unit-circle to describe the phase of the periodic trotting (fig. S23). Lastly, the target position is represented by a relative 3D vector with respect to the robot's local frame, and only the horizontal components are used as the state inputs. The detailed combination of the state observations for different tasks and networks are in table S5.

### Action space

The benchmark of DRL-based locomotion (*32, 55*) shows that a suitable configuration for the action space can yield better performance and faster learning due to the compliant interaction: a DRL agent provides joint references and an impedance mode for controlling the joint. This setting matches the equilibrium-point hypothesis for our proposed two-layer hierarchical control, and hence we adopted the same design of the action space here. To guarantee smooth and feasible actions for the real robot, we developed two important techniques: (i) the use of low-pass filters to emulate the characteristics of the frequency response of actuators, which enforces physically realisable reference motions, and (ii) the design of a particular loss function to generate smooth and non-jerky joint references and torques. We named these two techniques action filtering and smoothing loss, respectively.

### Action filtering

Real actuators have limited control bandwidth and hence the references with frequencies higher than the bandwidth cannot be tracked. However, a common issue in simulation is that the learning policy takes advantage of the pure torque source with unlimited control bandwidth: exploitation of abrupt and jerky motions that are only possible in simulation to maximise the reward but infeasible on real systems. The difference between ideal actuators (pure torque source) in simulation and real actuators (restricted bandwidth, torque, speed and power) needs to be addressed appropriately in the learning framework. In addition to the basic position and velocity limit (*56*), we performed action filtering using a first-order Butterworth filter to emulate the frequency response of real motors and to guide the policy to learn a smoother and more feasible behaviour.

For the *Jueying* robot, we found that emulating the frequency response and setting the speed limit are sufficient to represent realistic characteristics of actuators. The properties of *Jueying's* actuators, such as good torque tracking control and low gear ratio which results in decoupled inertia and minimal gear friction, avoid the need for modelling detailed actuator properties and simplify the simulation setting. During the simulation training, we emulated the limited frequency response of actuators by applying action filtering on the output action with the cut-off frequency of 5 Hz, which was higher than the 1.67 Hz trotting gait (fig. S24). This provides realistic restriction of high-frequency actions and prevents the policy from over-exploiting risky motions while still permitting necessary movements for dynamic tasks. For safety reasons in real experiments, we applied a more conservative cut-off frequency of 3 Hz in case of unexpected jerky references. As a result, all obtained policies exhibited smooth motions within the bandwidth (fig. S24) that can be executed on the real robot directly and safely.

*Smoothing loss*
The action filtering alone may not always guarantee feasible motions, because it only limits the frequency of the DNN output but not the magnitude, so the learning process may still explore and exploit low-frequency but large-amplitude motions regardless. Therefore, we further designed the smoothing loss based on the principle of minimal interaction to minimise the applied torques (*42*).

Biological studies show that when the CNS resets a new equilibrium, the displacement between the equilibrium and the actual position will activate a neuromuscular response that tries to reduce the muscular activity (torque). This principle of minimal interaction serves as a biological foundation of studying the proposed smoothing loss, which is effective for smoothing the exerted torque, i.e., $\tau = K_p(q^d - q^m)$. To guide the policy and generate actions following the minimal interaction principle, we designed a smoothing loss function $J_{smoothing}$ as:

$$J_{smoothing}(\mu(s_t)) = ||\mu(s_t) - q||_2, \qquad (1)$$

where $\mu(s_t)$ are the deterministic mean outputs of the stochastic policy used as the target joint references, and $q$ are the measured joint positions. The smoothing loss $J_{smoothing}$ is the objective function that minimises the differences between the target $\mu(s_t)$ and the measurement $q$. As the joint references are the inputs for impedance control, this minimisation leads to more gentle torque profiles, thus encouraging the learning of strategies with the least effort as possible.

The proposed smoothing loss $J_{smoothing}$ is incorporated into the SAC training loss $J_{SAC}(\pi)$ and is used for backpropagation of the neural networks, instead of being part of the reward function. Adding the smoothing loss term to $J_{SAC}(\pi)$ allows the information (the causality of actions) to backpropagate directly through the neural network and to bypass the process of reward bootstrap and Q-function approximation. Since for training the policy, the Q-function requires iterations to obtain a valid and accurate enough approximation of the expected return, our approach of bypassing the Q-function avoids the wait time and permits the information to backpropagate within the first few iterations.

**MELA training procedure**

Figure 6 depicts both the network architecture and training procedure of our MELA framework. The MELA network consists of one gating network and eight expert networks (Fig. 6B). The gating network has 2 hidden layers with 128 neurons each, using a ReLU activation function. All expert networks have 2 hidden layers with 256 neurons each and use a ReLU activation function, which has the same network structure as that of the pre-trained expert policy networks shown in Fig. 6A.

Here, we elaborate on MELA's two-stage training procedure. In stage 1, successful trotting and fall recovery policies were pre-trained using the scheme shown in Fig. 6A. In stage 2, MELA first initialised two groups of expert networks (four in each group) by copying the weights and bias from the two pre-trained experts (Fig. 6B) and randomly initialised the weights and bias of the gating network. Then MELA embedded all the experts together with the gating network, and co-trained all of them with diverse samples. During the co-training in stage 2, the gating network needed to learn how to compute correct weights for all experts and synthesise a new skill-adaptive network, as illustrated in Fig. 6B. The robot feedback states were the input of the synthesised network for generating motor actions.

Following the framework in Fig. 6, both stages were trained using the SAC algorithm (note S2), and the samples were collected at 25 Hz frequency while the actions were executed through the impedance control at 1000 Hz frequency. Both training stages initialised the robot in diverse configurations during the simulation episodes so as to increase the diversity of the collected samples, and additional details are described in note S2. The learning curves during stage1 and stage 2 of the MELA training can be found in fig. S25.

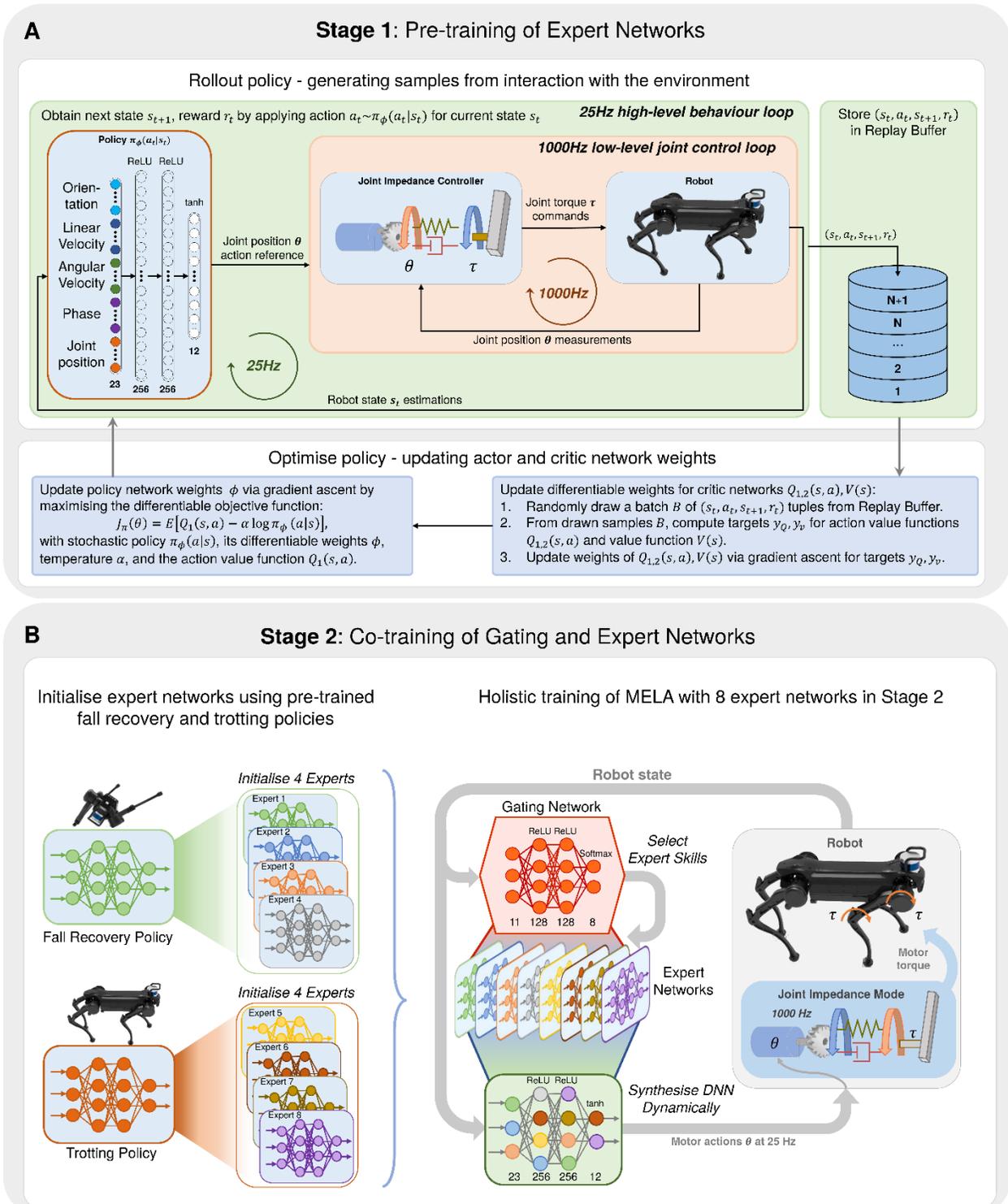

**Fig. 6. Two-stage training of MELA.** (**A**) In stage 1, the fall recovery and trotting policies are individually trained. (**B**) In stage 2, the pre-trained trotting and fall recovery policies from stage 1 are used to initialise two evenly distributed groups of experts, each containing 4 experts. All these expert networks are co-trained together with the gating network.

All neuron connections within MELA are differentiable, including those between the gating network and expert networks. This allows every network weight and bias to update through backpropagation simultaneously. Thus, all the MELA networks can be trained with the same backpropagation techniques used for Fully Connected Neural Networks. The *actor* for SAC was encoded using a MELA network, while the *critic* consisting of two Q functions was encoded as Fully Connected Neural Networks which were adopted from Double Q-learning to prevent overestimation (*57, 58*).

Let $x, y, h$ denote the dimensions of the input, the output and the hidden layer, respectively; let $W$ and $B$ be the network's weights and bias. The parameter-set of the skill-adaptive network is:

$$\psi_{synth} = \{W_0 \in \mathbb{R}^{x \times h}, W_1 \in \mathbb{R}^{h \times h}, W_2 \in \mathbb{R}^{h \times y}, B_0 \in \mathbb{R}^x, B_1 \in \mathbb{R}^h, B_2 \in \mathbb{R}^y\}, \qquad (2)$$

and the parameter-set of each individual expert network is:

$$\psi_{expert}^n = \{W_0^n \in \mathbb{R}^{x \times h}, W_1^n \in \mathbb{R}^{h \times h}, W_2^n \in \mathbb{R}^{h \times y}, B_0^n \in \mathbb{R}^x, B_1^n \in \mathbb{R}^h, B_2^n \in \mathbb{R}^y\}. \qquad (3)$$

During runtime, the weights $W$ and bias $B$ are fused by the weighted sum formulation as:

$$W_i = \sum_{n=1}^{8} \alpha_n W_i^n, B_i = \sum_{n=1}^{8} \alpha_n B_i^n, \qquad (4)$$

where $n = 1, \ldots, 8$ is the index of experts, $i = 0, 1, 2$ is the index of the corresponding layer, and $\alpha_n \in [0,1]$ are the variable weights generated by the gating network.

The fused weights $W$ and bias $B$ are used to construct the synthesised network dynamically during runtime using the equation as follows:

$$\Phi_{synth} = Tanh(W_2 ReLU(W_1 ReLU(W_0 X + B_0) + B_1) + B_2), \qquad (5)$$

where $X \in R^x$ is the input parameter, $Tanh(*)$ and $ReLU(*)$ are the nonlinear activation functions. The sum of the 8 variable weights $\alpha_n, (n = 1, \ldots, 8)$ is normalised to 1 using a Softmax function, also known as normalised exponential function. There are several nonlinear features in the blending process: each expert DNN is a nonlinear control policy by nature, and each blending of weight is produced by a nonlinear rescaling of the output of the gating network with a Softmax function, which normalises different values of the original sum. Therefore, the resulting synthesised expert is a highly nonlinear control policy – a nonlinear mapping between the feedback states and actions that is required to deal with challenging scenarios.

**SUPPLEMENTARY MATERIALS**

Note S1. Data analysis of MELA learning results.
Note S2. Additional Materials and Methods.
Note S3. Expert imbalance phenomenon.
Fig. S1. Comparison of MELA's learning curves using different numbers of expert networks.
Fig. S2. Baseline experiments of fall recovery and trotting from engineered controllers.
Fig. S3. Activation patterns of experts across all motor skills.
Fig. S4. Five representative cases showing adaptive behaviours of the MELA expert under unseen situations in simulation.
Fig. S5. Forward trotting velocity during the variable-speed trotting simulation.
Fig. S6. Heading angle and angular velocity during the steering experiment on the real robot.
Fig. S7. Relative target positions with respect to the robot from the user command as the input to the MELA networks during the real locomotion experiment.
Fig. S8. Measured torques of the front left leg during the real locomotion experiment (Fig. 5D).
Fig. S9. Roll and pitch angles during the real locomotion experiment.
Fig. S10. Target position, body orientation and velocity during the real locomotion experiment on grass (Fig. 5E).
Fig. S11. Adaptive and agile locomotion behaviours on pebbles and grass demonstrated by MELA policy.
Fig. S12. Continuous and variable weights of all experts during the real MELA experiment (Fig. 5D).
Fig. S13. Continuous and variable weights of all experts during the real MELA experiment on grass (Fig. 5E).
Fig. S14. Four types of unseen terrains for testing the multi-skill MELA policy in the simulation.
Fig. S15. Simulated test scenarios for evaluating the robustness of the MELA policy.
Fig. S16. Representative adaptive behaviour from the simulated scenario of steering on spot (fig. S4A, movie S6).
Fig. S17. Representative adaptive behaviour from the simulated scenario of steering while recovering to trotting (fig. S4B, movie S6).
Fig. S18. Representative adaptive behaviour from the simulated scenario of tripping (fig. S4C, movie S6).
Fig. S19. Representative adaptive behaviour from the simulated scenario of a large impact (fig. S4D, movie S6).
Fig. S20. Representative adaptive behaviour from the simulated scenario of falling off a cliff (fig. S4E, movie S6).
Fig. S21. Analysis of responses from the MELA policy during the simulated scenario of a large external perturbation (fig. S4D, movie S6).
Fig. S22. Phenomenon of asymmetric gait and imbalanced experts from MoE.
Fig. S23. Illustration of the 2D phase vector for training the locomotion policy.

Fig. S24. Normalised power spectrum analysis of motions during the real locomotion experiment (without the DC component).
Fig. S25. Learning curves during the 2-stage training of MELA and MoE.
Fig. S26. Nine distinct configurations used as the initialisation for training fall recovery policies in simulation.
Fig. S27. Setting of the target location for training MELA policies in simulation.
Table S1. Distribution matrix of expert specialisations over motor skills.
Table S2. Specification of the Jueying quadruped robot.
Table S3. Detailed descriptions of the individual reward terms.
Table S4. Weights of the reward terms for different tasks.
Table S5. Selection of state inputs for different tasks and neural networks.
Table S6. Proportional-Derivative parameters for the joint-level PD controller.
Table S7. Hyperparameters for SAC.
Movie S1. Experiments of fall recovery.
Movie S2. Experiments of outdoor fall recovery and compliant interactions.
Movie S3. Experiments of trotting.
Movie S4. Experiments of adaptive locomotion and behaviours.
Movie S5. Experiments of outdoor fall-resilient locomotion on irregular terrains.
Movie S6. Simulation of representative adaptive behaviours from the multi-skill MELA expert.
Movie S7. Simulation of extensive scenarios and crash tests.
Movie S8. Baseline experiments of fall recovery and trotting from default controllers.

**Acknowledgments**
The authors have special thanks to Prof. João Ramos and Dr. Kseniya Korobchevskaya who read an early draft of this manuscript and offered valuable suggestions, and Christopher McGreavy for proofreading this manuscript and producing the artwork. The corresponding author has special thanks to Dr Albert Mukovskiy for his guidance on the equilibrium-point theories since the AMARSi project. Sincere appreciation to DeepRobotics Co. Ltd for permitting all crash tests, and those who helped with experiments: Zhaole Sun, Jiangyuan Zhang, Xiaobo Mo, Chengxiao Li, Zhen Chu, and Chao Li. **Funding**: This research work is partly supported by the EPSRC CDT in Robotics and Autonomous Systems (EP/L016834/1), the Open Project (ICT1900349, ICT20005) from Zhejiang University, and the rest is self-financed. **Author contributions**: C.Y. contributed to the multi-expert learning structure, simulation, data analysis, experiments, and the manuscript. K.Y. contributed to the hierarchical learning structure, C++ software on the robot, simulation, experiments, data collection, and the manuscript. Q.Z. contributed to the development of Jueying robots, hardware facilities and supported all robot experiments. W.Y. contributed to the writing, figures, and data analysis in the manuscript. Z.L. directed the research, designed and debugged key hardware experiments, and authored the manuscript. **Competing Interests**: The authors declare that they have no competing interests. **Data and materials availability**: All data needed to evaluate the conclusions in the paper are present in the paper or the Supplementary Materials.

# Science Robotics

AAAS

# Supplementary Materials for

## Multi-expert learning of adaptive legged locomotion

Chuanyu Yang*, Kai Yuan*, Qiuguo Zhu, Wanming Yu, Zhibin Li†

†Corresponding author: zhibin.li@ed.ac.uk

**This PDF file includes:**

Note S1. Data analysis of MELA learning results.
Note S2. Additional Materials and Methods.
Note S3. Expert imbalance phenomenon.
Fig. S1. Comparison of MELA's learning curves using different numbers of expert networks.
Fig. S2. Baseline experiments of fall recovery and trotting from engineered controllers.
Fig. S3. Activation patterns of experts across all motor skills.
Fig. S4. Five representative cases showing adaptive behaviours of the MELA expert under unseen situations in simulation.
Fig. S5. Forward trotting velocity during the variable-speed trotting simulation.
Fig. S6. Heading angle and angular velocity during the steering experiment on the real robot.
Fig. S7. Relative target positions with respect to the robot from the user command as the input to the MELA networks during the real locomotion experiment.
Fig. S8. Measured torques of the front left leg during the real locomotion experiment (Fig. 5D).
Fig. S9. Roll and pitch angles during the real locomotion experiment.
Fig. S10. Target position, body orientation and velocity during the real locomotion experiment on grass (Fig. 5E).
Fig. S11. Adaptive and agile locomotion behaviours on pebbles and grass demonstrated by MELA policy.
Fig. S12. Continuous and variable weights of all experts during the real MELA experiment (Fig. 5D).
Fig. S13. Continuous and variable weights of all experts during the real MELA experiment on grass (Fig. 5E).
Fig. S14. Four types of unseen terrains for testing the multi-skill MELA policy in the simulation.
Fig. S15. Simulated test scenarios for evaluating the robustness of the MELA policy.
Fig. S16. Representative adaptive behaviour from the simulated scenario of steering on spot (fig. S4A, movie S6).
Fig. S17. Representative adaptive behaviour from the simulated scenario of steering while recovering to trotting (fig. S4B, movie S6).
Fig. S18. Representative adaptive behaviour from the simulated scenario of tripping (fig. S4C, movie S6).
Fig. S19. Representative adaptive behaviour from the simulated scenario of a large impact (fig. S4D, movie S6).
Fig. S20. Representative adaptive behaviour from the simulated scenario of falling off a cliff (fig. S4E, movie S6).
Fig. S21. Analysis of responses from the MELA policy during the simulated scenario of a large external perturbation (fig. S4D, movie S6).
Fig. S22. Phenomenon of asymmetric gait and imbalanced experts from MoE.
Fig. S23. Illustration of the 2D phase vector for training the locomotion policy.
Fig. S24. Normalised power spectrum analysis of motions during the real locomotion experiment (without the DC component).
Fig. S25. Learning curves during the 2-stage training of MELA and MoE.
Fig. S26. Nine distinct configurations used as the initialisation for training fall recovery policies in simulation.
Fig. S27. Setting of the target location for training MELA policies in simulation.
Table S1. Distribution matrix of expert specialisations over motor skills.
Table S2. Specification of the Jueying quadruped robot.
Table S3. Detailed descriptions of the individual reward terms.
Table S4. Weights of the reward terms for different tasks.
Table S5. Selection of state inputs for different tasks and neural networks.
Table S6. Proportional-Derivative parameters for the joint-level PD controller.
Table S7. Hyperparameters for SAC.

**Other supplementary files for this manuscript include:**

Movie S1. Experiments of fall recovery.
Movie S2. Experiments of outdoor fall recovery and compliant interactions.
Movie S3. Experiments of trotting.
Movie S4. Experiments of adaptive locomotion and behaviours.
Movie S5. Experiments of outdoor fall-resilient locomotion on irregular terrains.
Movie S6. Simulation of representative adaptive behaviours from the multi-skill MELA expert.
Movie S7. Simulation of extensive scenarios and crash tests.
Movie S8. Baseline experiments of fall recovery and trotting from default controllers.

# Notes

**Note S1. Data analysis of MELA learning results**
From fig. S12A, the changing weights around the boundaries of locomotion modes indicate that MELA produced smooth and quick transitions across successive modes. The data in fig. S12C-E shows a clear correlation between the sum of the experts' weights and the locomotion mode, suggesting that MELA trained the experts to be activated in a collaborative manner. Figure S12C delineates the sum of the weights of expert 3 and 7, which is high throughout the trotting mode but low during standing and fall recovery. Similar patterns of activation can also be observed in fig. S12D, where the sum of expert 5, 6 and 8 has particularly high peaks during fall recoveries but constantly low in other modes. Figure S12E shows the differential weight between expert 1 and 4 (weight of expert 1 deducted by that of expert 4), and the complementary activation during left and right trotting, i.e., activation of expert 1 and inhibition of expert 4 during right trotting, and vice versa.

In conclusion, the data in fig. S12 reveals the correlations and patterns between the weights of collaborating experts and the corresponding locomotion mode, suggesting that trained experts in the MELA were activated in a coherent manner.

**Note S2. Additional Materials and Methods**
In this section, we provide additional details of the training procedures and the control framework that are needed to reproduce our presented policies.

**Sample collection procedure**
The distribution of samples collected by the agent during training will affect the learning outcome of the policy. In order to obtain a sample distribution containing a variety of robot states for better generalisation, we used two techniques to augment the standard sample collection: reference state initialisation and early termination.

First, to increase the diversity among collected samples, we initialised each simulation episode by a random selection from a set of reference configurations. The diversity in the sample distribution allows the agent to learn a policy of good generalisability for a range of different states. Furthermore, by initialising the robot in difficult configurations, the agent can experience challenging cases more often. Second, we applied early termination to the episode during training when the robot encountered an undesirable state, such as a failure state in which the robot was unable to recover. Early termination prevents irreversible failures from skewing the sample distribution (note S2).

**Control framework**
Joint torque control of the real *Jueying* robot is used to create an impedance mode for all joints because having mechanical impedance is known to be robust during the physical contact and interactions (*59*, *60*). Hence, the synthesised DNN plays a role similar to a CNS that produces trajectory attractors constantly pulling and pushing all joints in the impedance mode to generate joint torques similar to a spring-damper system.

The neural network policy generates joint position references at 25 Hz (Fig. 6A). A policy with an update frequency of around 25 Hz is a common setting as a high-level motion planning layer in robot control (*27, 28, 31, 61*). The standard joint-level trajectory interpolation and speed limit were implemented to generate smooth position references at 1000 Hz for the low-level impedance control.

The Proportional-Derivative (PD) gains used for the impedance control are shown in table S6. The feedback gains were 700Nm/rad and 10Nms/rad for stiffness and damping respectively and were sufficient for the robot to have good control over its swing leg and placement of stance foot. Based on our proposed smoothing loss, the policy has learned an active compliance behaviour in the simulation as well as in the experiments by adjusting the position references. The supplementary video (movie S2) shows that the robot interacted in a compliant manner, instead of being stiff as it would be in a pure position control mode with the same PD gains.

During the dynamic response, the stiffness produced by the active control can be lower than the original PD gains set in the impedance mode, as it is regulated the same way as for series elastic actuators (SEA) (*48*). Therefore, it is desirable to have medium PD gains in the low-level joint control, and render the desired or a lower impedance by adjusting the high-level position set-point, i.e., a deliberate motion to buffer an impact.

The MELA policy learns how to regulate the set-point for the impedance controller to achieve compliant behaviour, which is similar to active compliance found in the control of robotic arms. Robot manipulators are usually controlled by high PD gains and are thus very stiff, but soft and compliant behaviours can still be achieved by limiting the amount of torque applied on joints (*62, 63*). Since the neural network receives feedback of the actual joint positions $q^m$ and has direct control of the desired joint positions $q^d$, according to the SEA principle $\tau = K_p(q^d - q^m)$, actively changing the set-point $q^d$ with respect to the measured position $q^m$ can restrict the amount of torque, lower the stiffness, and thus increase the compliance.

**Nomenclature**

This note describes the definitions of mathematical notations to help explain the equations of the reward terms in the following section.

| | |
|---|---|
| $\phi^{base}$ | Orientation vector: the projection of the normalised gravity vector in the robot base frame to represent the orientation |
| $h^{world}$ | The robot base height $(z)$ in the world frame |
| $v_{base}^{world}$ | The linear velocity of the robot base in the world frame |
| $v_{base}^{local}$ | The linear velocity of the robot base in the robot's local heading frame: $R^T(\theta_{yaw}^{world}) \times v_{base}^{world}$ |
| $\theta_{yaw}^{world}$ | The yaw orientation of the robot body in the world frame |
| $\omega_{yaw}^{world}$ | The yaw angular velocity of the robot body in the world frame |
| $\tau$ | The vector of all joint torques |
| $q$ | The vector of all joint position |
| $\dot{q}$ | The vector of all joint velocity |
| $\widehat{(\cdot)}$ | The desired quantity of selected property $(\cdot)$, where $(\cdot)$ is a placeholder |
| $h_{foot,n}^{world}$ | The n-th foot height $(z)$ in the world frame |
| $v_{foot,n}^{world}$ | The horizontal linear velocity $(\dot{x}, \dot{y})$ of the n-th foot in the world frame |
| $p_{goal}^{world}$ | The horizontal component of the goal position $(x, y)$ in the world frame |
| $p_{robot}^{world}$ | The horizontal component of the robot base position in the world frame |
| $u_{goal,base}^{base}$ | The unit vector pointing from the robot base to the goal at the base frame |
| $p_{foot,n}^{world}$ | The horizontal placement of the n-th foot in the world frame |

**Soft Actor Critic**

Compared to the classic reinforcement learning methods that obtain a policy $\pi(s_t)$ by maximising the expected sum of rewards as:

$$J(\pi) = \sum_{t=0}^{T} E_{(s_t, a_t) \sim \rho_\pi}(r(s_t, a_t)), \qquad (1)$$

we used the Soft Actor Critic (SAC) with the objective $J(\pi)$ optimised over an additional maximum entropy objective $H(\pi(\cdot | s_t))$ as:

$$J_{SAC}(\pi) = \sum_{t=0}^{T} E_{(s_t, a_t) \sim \rho_\pi}(r(s_t, a_t) + \alpha H(\pi(\cdot | s_t))), \qquad (2)$$

where $H(\pi(\cdot | s_t))$ is the expected entropy of policy $\pi$ over the sample distribution $\rho_\pi$. The temperature parameter $\alpha$ affects the stochasticity and the exploration capability of the optimal policy by changing the influence of the entropy term, where a higher $\alpha$ leads to more stochastic policies, and vice versa. SAC balances the well-known problem of exploitation and exploration by automatically tuning the temperature parameter $\alpha$.

The policy is expressed as a Gaussian distribution $\mathcal{N}(\mu_\psi(s_t), \sigma_\psi(s_t)^2)$ with mean $\mu_\psi(s_t)$ and covariance $\sigma_\psi(s_t)$ generated by a neural network. A *tanh* function is applied to the Gaussian samples to project the action distribution within $\hat{a}(s_t) \in (-1, 1)$. The squashed action $\hat{a}(s_t)$ is then scaled by the joint range $\bar{q}$ to formulate the target action $a(s_t) = \bar{q}\hat{a}(s_t)$ as the joint position references for the robot. The hyperparameters of the SAC algorithm can be found in table S7.

**Initialisation and early termination**

We designed a set of initial reference states to initialise episodes for each task. As shown in fig. S26, for fall recovery, we designed 9 distinct poses to ensure the diversity within the collected samples: (i) standing at nominal height, (ii) standing at maximal height with straight knees, (iii) leg sprawling posture, (iv) lying on the back, (v) lying on the left side, (vi) lying on the right side, (vii) crouching, (viii) kneeling, and (ix) lying on the abdomen. Posture (i) - (vi) are common configurations during standing and fall recovery, while posture (vii) - (ix) are unusual contact configurations that are difficult to recover. For learning to trot, a trotting gait sample from the robot's factory setting was used as reference states for imitation. We combined the reference state initialisation datasets of fall recovery and trotting to create a new set of initialisation states for the multimodal locomotion, since the target-following using MELA would involve all modes and their transitions. For training the MELA policy, the goal position was initialised within a circular area of 6 m radius around the robot at the beginning of each episode (fig. S27).

We specified three termination criteria for locomotion related tasks: (i) any body part other than feet are in contact with the ground, (ii) the body orientation exceeding a threshold of 90°, and (iii) reaching the time limit of the episode. For the fall recovery task, only the 3rd criterion was used because the robot had to undergo the states specified in criteria (i) and (ii) to learn how to recover from failures autonomously.

**Reward design**
We used a radial basis function (RBF) to design a bounded reward function:

$$\varphi(x,\hat{x},\alpha) = exp(\alpha(\hat{x}-x)^2) , \qquad (3)$$

where $x$ is the physical quantity for the evaluation, $\hat{x}$ is the desired value, and $\alpha$ is the parameter that controls the width of the RBF. We utilised the RBF for all reward terms that involve continuous physical quantities. Reward terms with discrete properties, i.e., foot-ground contact, body-ground contact, and ground contact imitation, are designed separately. Details of the individual reward terms using the aforementioned nomenclature are presented in table S3 and table S4.

In table S3, the first twelve terms are straightforward and task-related since each evaluates a physical quantity directly related to the physical movements. The *reference joint position reward* and *reference foot contact reward* provide reference trajectories from an existing trotting gait for the agent to imitate. By providing such a reference gait, the agent is able to learn stable trotting more efficiently and effectively through imitation. We also clarify the purpose of the last two remaining reward terms in table S3 as follows. The *Swing and stance reward* reflects the contact constraint that encourages a higher velocity at a smaller height error to encourage the swing motion, and a lower velocity at a larger height error while the feet deviate from the nominal height $\hat{h}$. By discouraging the stance foot to move, the *Swing and stance reward* is able to prevent slippage as well. The *foot placement reward* encourages the feet to place around the robot body averagely, guiding the policy to perform more symmetric and stable foot placement during locomotion.

**Note S3. Expert imbalance phenomenon**
In this section, we provide additional details of the implementation of MoE and its qualitative comparison with MELA. The MoE approach synthesises multiple experts using the weighted sum of the output of each expert network as:

$$o = \sum_{n=1}^{8} \alpha_n o_n, \qquad (4)$$

where $n = 1, ... , 8$ is the index of experts, $o_n$ are the output generated by the $n$th expert network, $\alpha_n \in [0, 1]$ are the variable weights generated by the gating network, and $o$ is the final synthesised output of MoE.

The MoE policy was obtained using the same pre-trained experts and training procedure as MELA. The comparison of the learning curves of MELA and MoE are presented in fig. S25B, where MELA shows a faster learning curve than MoE, e.g., a higher reward than MoE during 120-300 episodes. The training continued until 600 episodes to ensure that each policy had fully converged in order to be used for the comparison study.

The MoE and MELA policies were benchmarked in the same target-following test involving fall recovery, simultaneous trotting and steering. The target ball within the test scenario was designed to move forward in a straight line at a constant speed during 0-7s and then with a sinusoidal lateral motion superimposed on the forward motion during 7-40s.

We found that for fall recovery and forward trotting tasks, MoE and MELA policies show similar performance in these trained scenarios, because of the use of the same pre-trained experts and training settings. However, the performance becomes different when it comes to the adaptive behaviours that are represented by the newly emerged behaviours and transitional skills, which were not pre-trained: MELA steered left and right actively and followed the moving target very well, whereas MoE could not walk straight towards the target and its body was not oriented towards the direction of travel.

As shown in fig. S22A-C, the phenomenon of an asymmetric gait was observed from the MoE policy, which exhibits unbalanced left and right steering skills in contrast to the symmetric steering behaviour of the MELA policy, indicating that MoE's left and right steering experts are not trained well, with some experts favored more than others. During the target-following test, the robot controlled by MELA was able to turn left and right equally well for changing the heading direction progressively with the moving target, as shown in fig. S22A. However, the robot controlled by MoE could not follow the heading to the ball and was side stepping forward with unresponsive right steering, resulting in a constant offset of the heading and body yaw angle, as shown in the 3rd-5th snapshots in fig. S22B.

Figure S22C illustrates the global position of the target and the robot as well as the yaw angle using MELA and MoE, respectively. From 0-7s, the MELA-controlled robot trotted forward after getting up from a prone pose and tracked the zero yaw angle, while the MoE-controlled robot moved forward with an offset yaw angle because of degenerated steering experts. Moreover, during 7-40s, MELA showed equally active experts for left and right steering with a symmetric profile of yaw angle around zero, whereas MoE exhibits a drifting yaw angle and offset body pose due to unbalanced steering skills.

The t-SNE analysis in fig. S22D-E further reveals the underlying differences between MoE and MELA in fall recovery and trotting tasks respectively. In contrast to the distinct clusters from MELA (Fig. 4D-E), the lack of learning adaptive skills was observed in MoE that only some favored experts were well trained, leading to limited diversification of skills. In fig. S22D, only five experts are distinctly clustered (expert 1, 2, 5, 7, 8), while others are sparsely overlapped with each other. In fig. S22E, only four experts have individual clusters (expert 2, 5, 7, 8), while others are scattered. This t-SNE analysis shows that approximately half of the experts are imbalanced in MoE leading to under-trained skills, and such limited diversification is disadvantageous in learning adaptive motor skills.

**Figures**

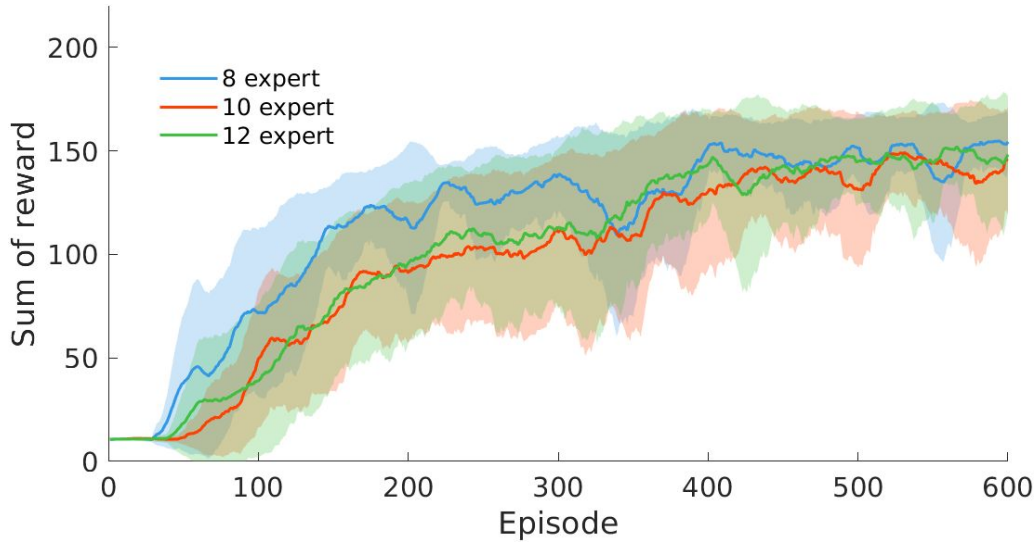

**Fig. S1. Comparison of MELA's learning curves using different numbers of expert networks**. It can be seen that using more than 8 experts does not improve the task performance, and has a slower convergence instead.

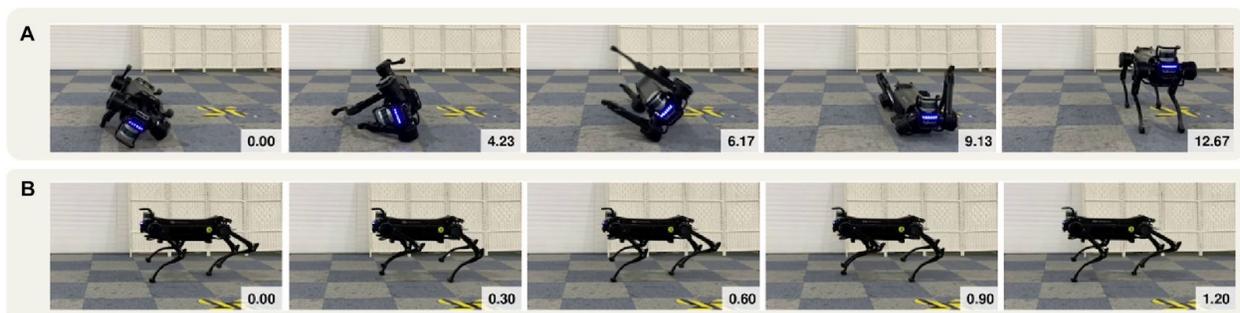

**Fig. S2. Baseline experiments of fall recovery and trotting from engineered controllers.** (**A**) The fall recovery from the engineered controller has a fixed sequence of motions and takes more than 12 s to stand up. (**B**) The trotting gait from the robot's control suite provided by the manufacturer. See movie S8 for more details.

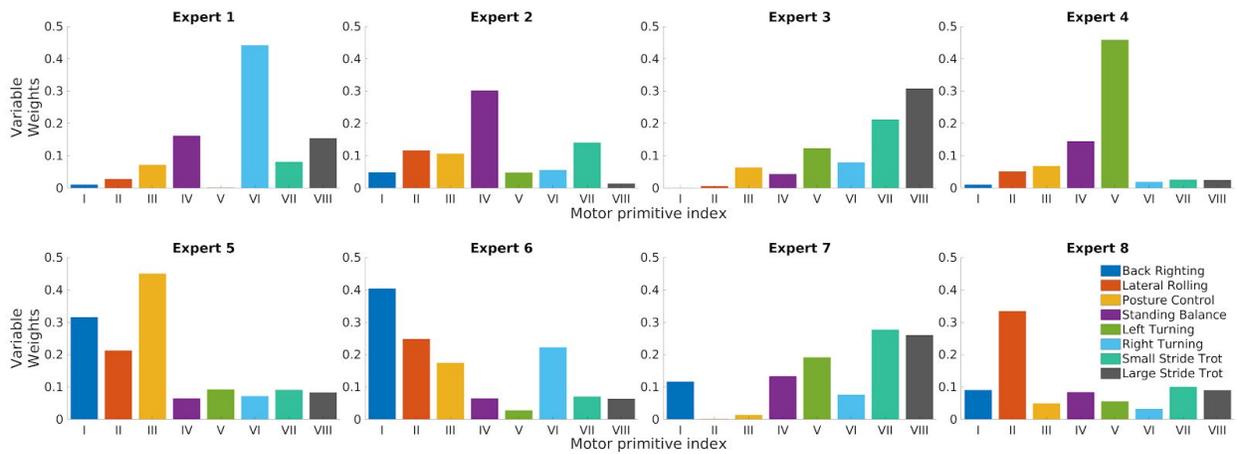

**Fig. S3. Activation patterns of experts across all motor skills.** The unique activation pattern of each expert, in which the specialisation is indicated by the highest activation of a motor skill numerated by roman numbers. The specialised motor skills of expert 1-8 are: (i) right turning, (ii) balance stabilisation, (iii) large-step trotting, (iv) left turning, (v) posture control, (vi) back righting, (vii) small-step trotting, and (viii) lateral rolling, respectively. The data used for visualising the activation patterns are obtained from simulation tests of the trained MELA policy.

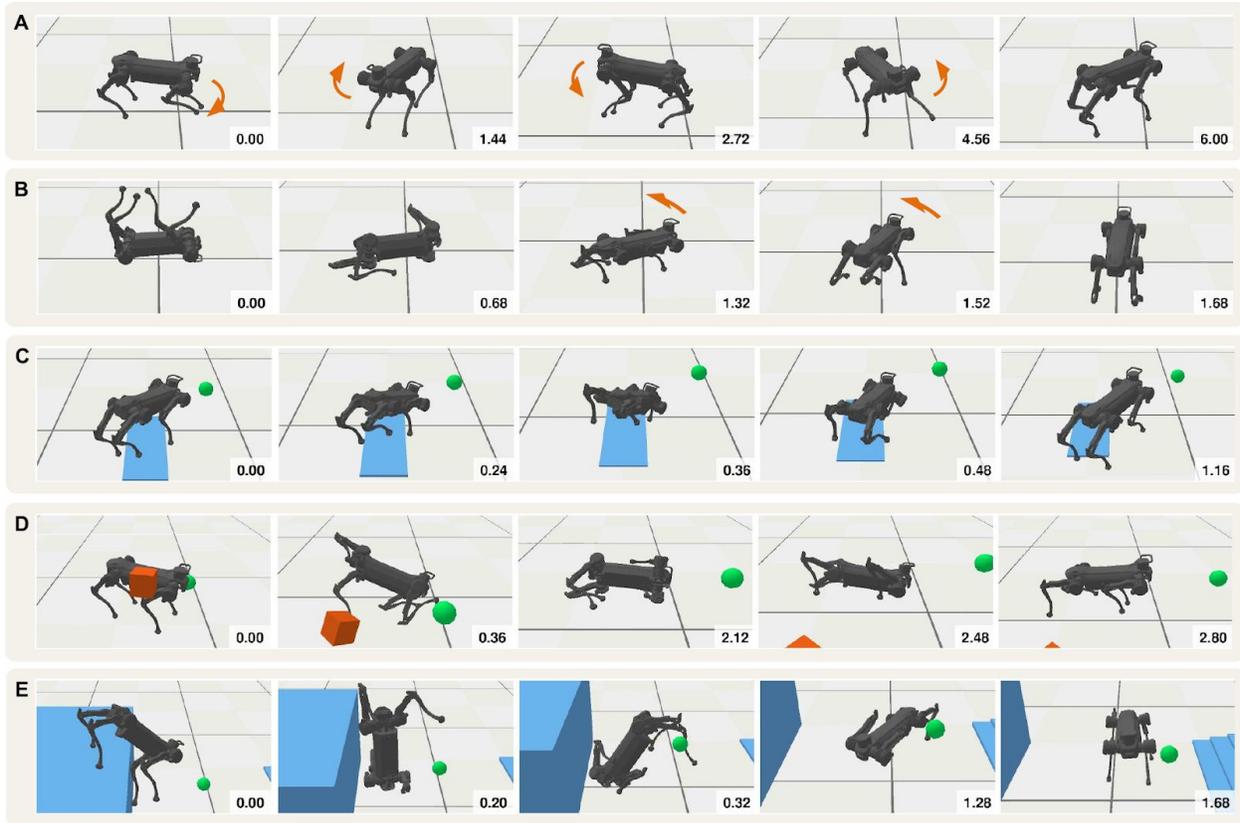

**Fig. S4. Five representative cases showing adaptive behaviours of the MELA expert under unseen situations in simulation.** (**A**) An emerged behaviour of left and right steering on the spot. (**B**) An emerged behaviour of simultaneous steering and standing up while recovering from fall to trotting. (**C**) A tripping case caused by a slippery ground with a low friction coefficient of 0.1. The tripping and recovery behaviour was similar to that in the real experiment (see Fig. 5D). (**D**) A large impact disturbance caused by a 20 kg box hitting the robot at 8 m/s velocity. (**E**) An extreme crash test of blind locomotion over a cliff of 1 m height. (Time in snapshots is in second).

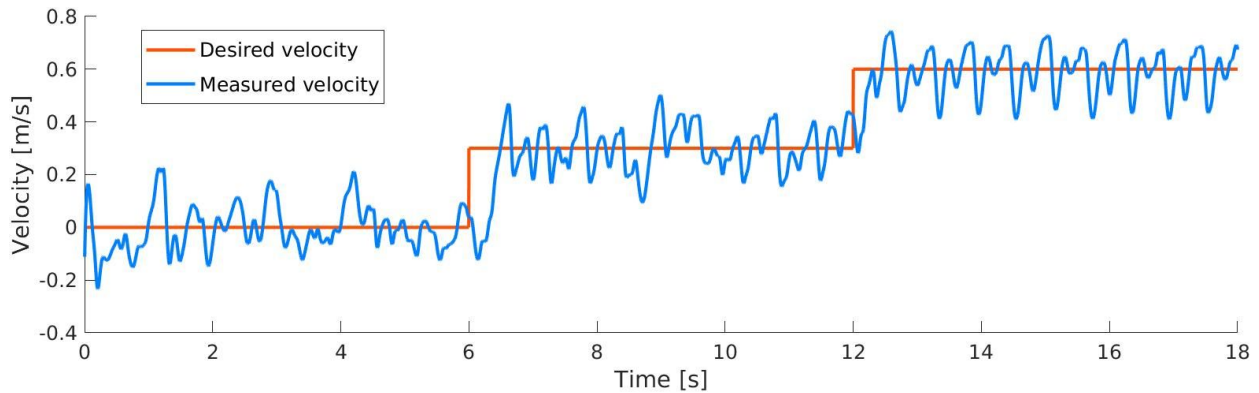

**Fig. S5. Forward trotting velocity during the variable-speed trotting simulation.** The robot adapted its trotting speed and followed the moving target (see movie S6).

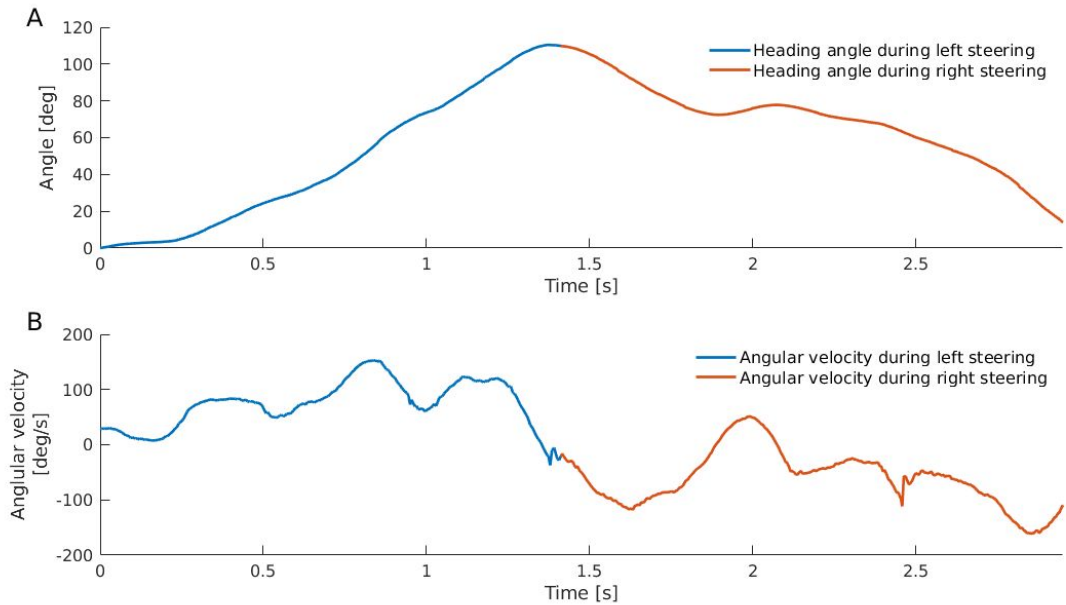

**Fig. S6. Heading angle and angular velocity during the steering experiment on the real robot.** The heading data here corresponds to the experiment in Fig. 5B. (**A**) The robot first steered counter-clockwise towards the left and then clockwise towards the right. (**B**) The average yawing velocities were 1.6 *rad/s* (92.0 *deg/s*) and -1.1 *rad/s* (-61.7 *deg/s*) during left and right steering, respectively, while the peak velocities reached 2.7 *rad/s* (156.8 *deg/s*) and -2.7 *rad/s* (-156.9 *deg/s*).

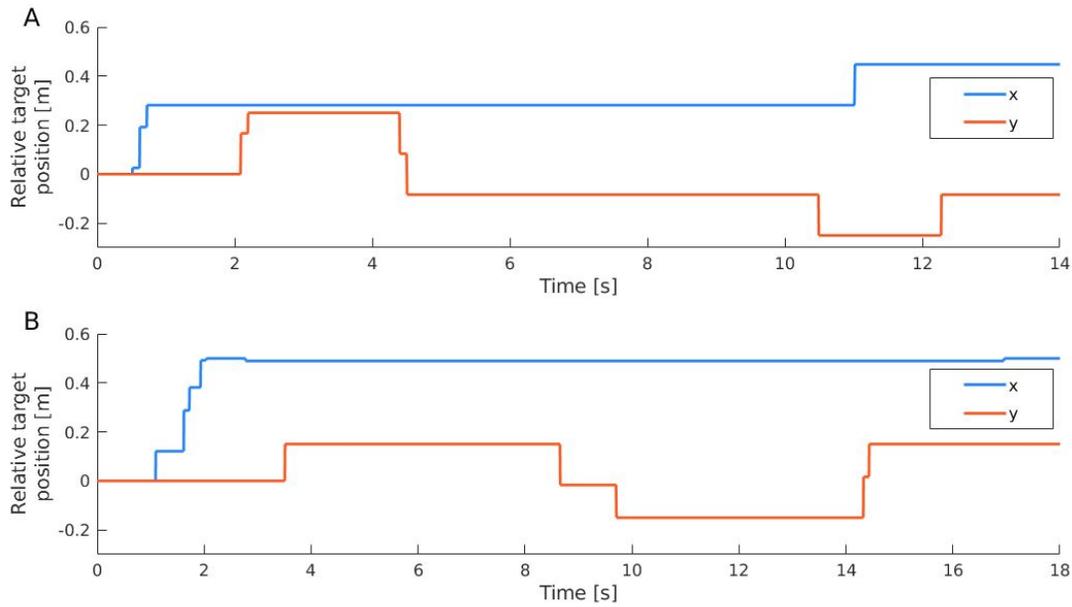

**Fig. S7. Relative target positions with respect to the robot from the user command as the input to the MELA networks during the real locomotion experiment.** (**A**) The changing target position (x, y) during the real target-following experiment (Fig. 5C), and runtime was 14 seconds. (**B**) The changing target position (x, y) during the fall-resilient experiment (Fig. 5D), and runtime was 18 seconds.

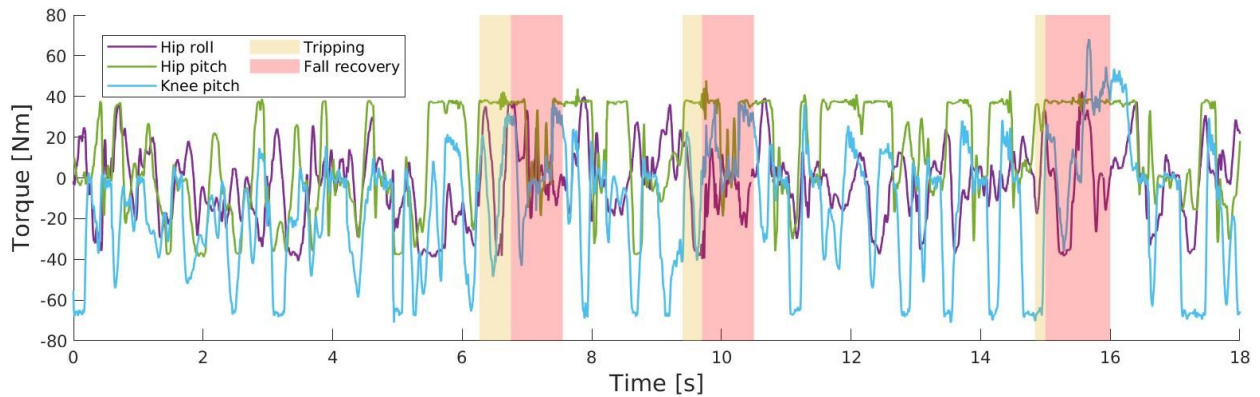

**Fig. S8. Measured torques of the front left leg during the real locomotion experiment (Fig. 5D).** During this experiment, the robot trotted at large steps (see the larger commanded (x, y) target positions in fig. S7B) and saturated the motor torques at times, e.g., the hip pitch joint. In this case, the torque-saturated leg was not able to move as intended and the robot stumbled and tripped (yellow regions), leading to fall recoveries (red regions).

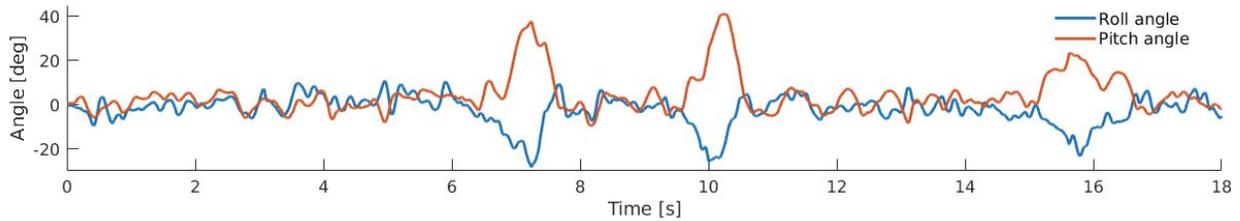

**Fig. S9. Roll and pitch angles during the real locomotion experiment.** The measurements correspond to the experiment presented in Fig. 5D. Significant changes were observed in the body orientation by the roll and pitch angles during the tripping moments. The peaks in roll and pitch angles were up to 0.47 *rad* (26.7 *deg*) and 0.7 *rad* (39.8 *deg*), respectively. The short duration of the large deviations of the body posture shows that the recovery was accomplished within 1 second time.

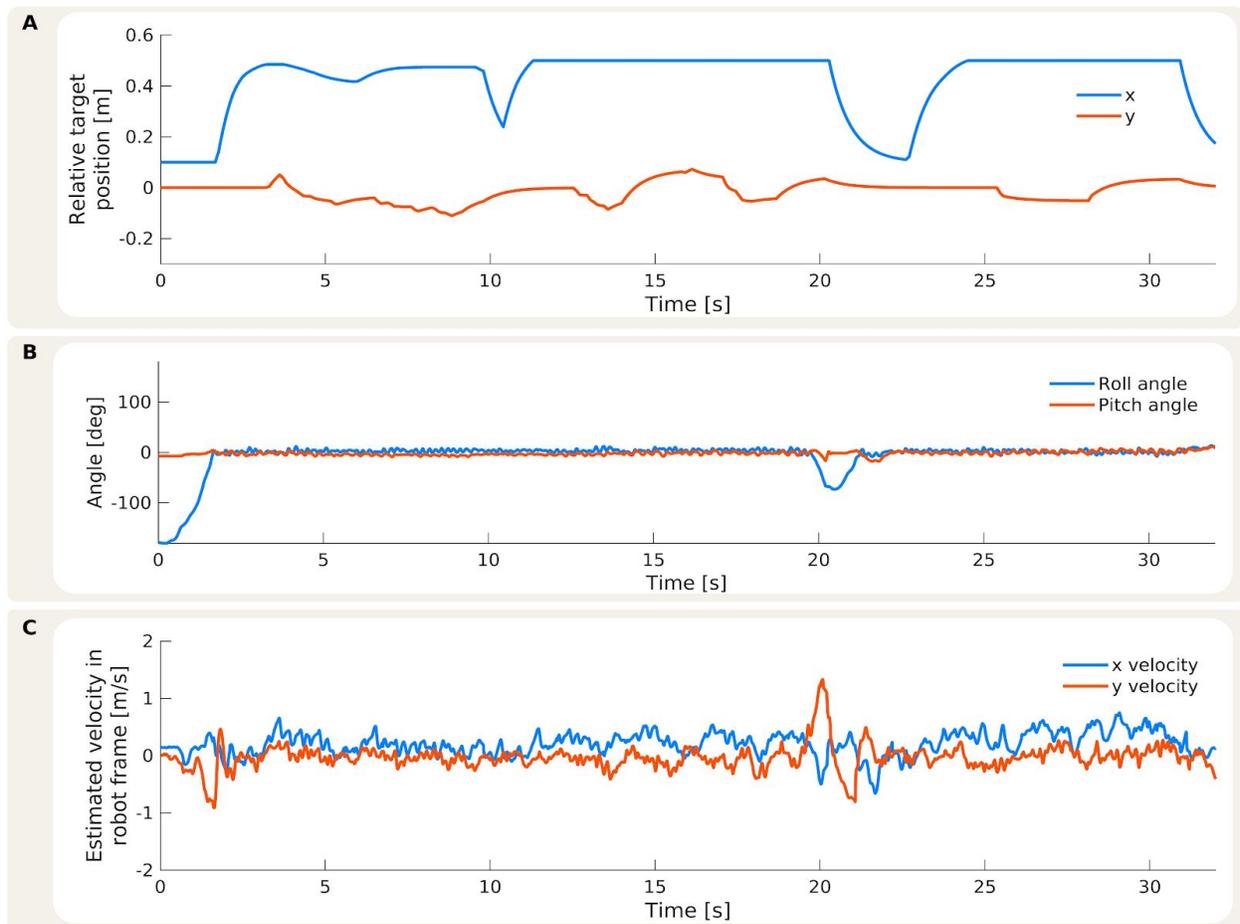

**Fig. S10. Target position, body orientation and velocity during the real locomotion experiment on grass (Fig. 5E). (A)** Relative target positions with respect to the robot from the user command as the input to the MELA networks. **(B)** The measured roll and pitch angles of the robot. The robot is able to maintain a steady posture while trotting on uneven grass. **(C)** The estimated velocity of the robot represented in the local robot frame. The robot trots with a forward velocity around 0.25 *m/s*. The robot experiences large lateral velocity at around the 20s moment when it is being knocked over.

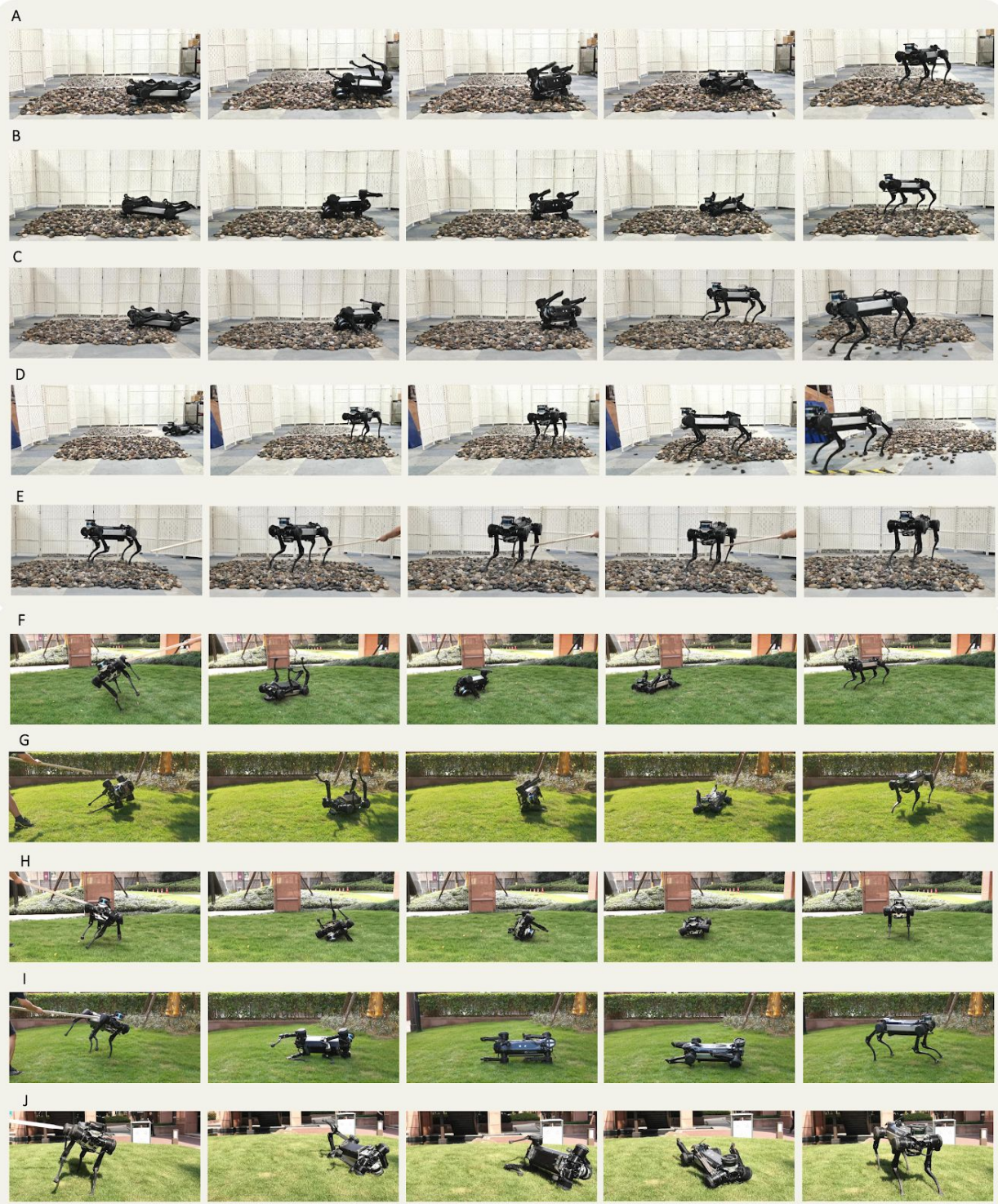

**Fig. S11. Adaptive and agile locomotion behaviours on pebbles and grass demonstrated by MELA policy.** (**A-E**) Self-righting and trotting on pebbles. (**F-J**) Self-righting, trotting, and adaptive fall recovery in presence of various pushes on grass.

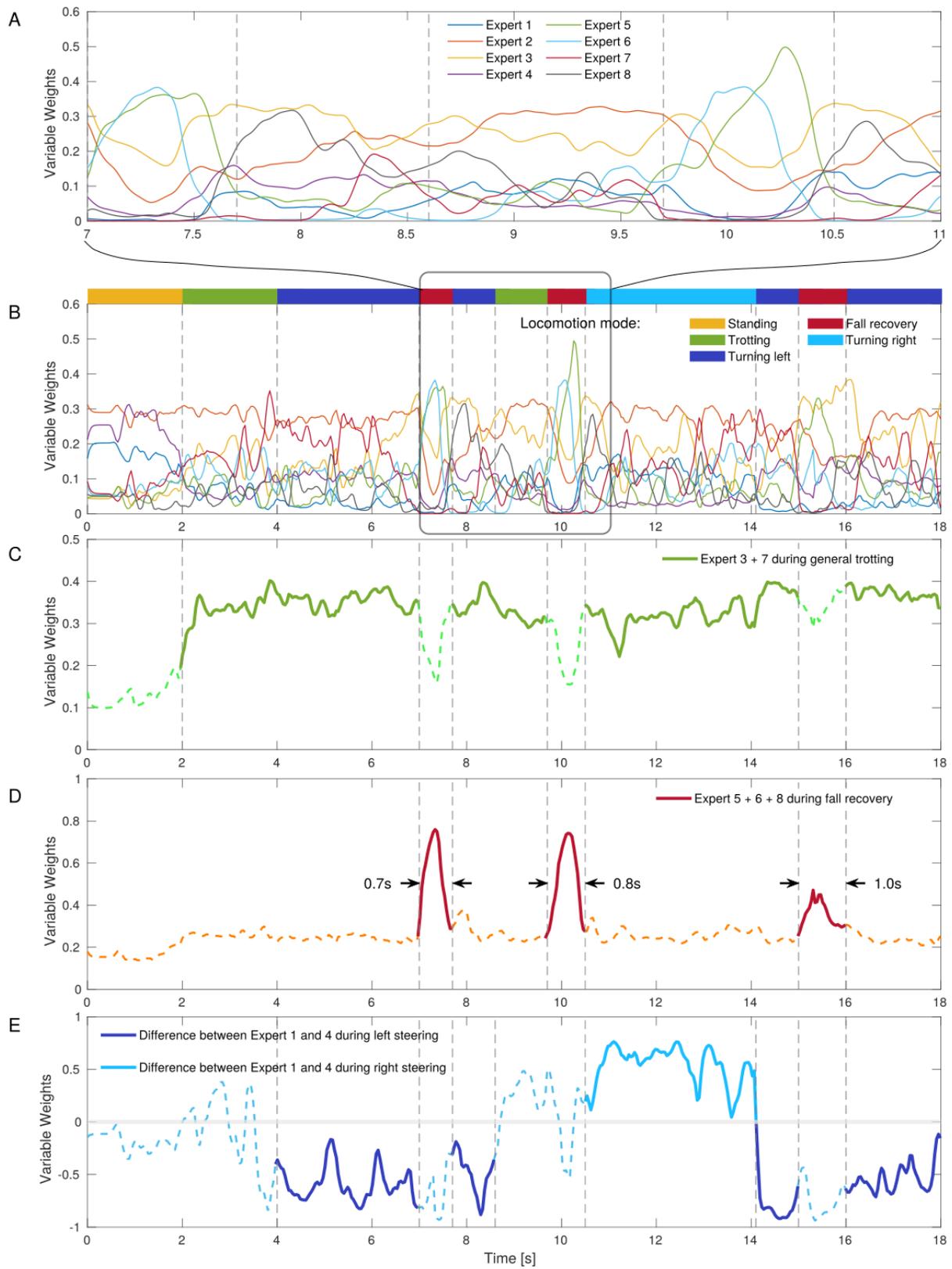

**Fig. S12. Continuous and variable weights of all experts during the real MELA**

**experiment (Fig. 5D).** (**A**) The variable activations within a zoomed period to show the transition of weights between multiple experts. (**B**) The variable activations of the entire multimodal locomotion with trotting, turning and fall recovery during a target-following task. (**C-E**) The activation levels of paired weights from collaborating experts, where the expert groups (3, 7), (5, 6, 8), (1, 4) cooperated together in trotting (forward, left, right), fall recovery, and turning (left, right), respectively.

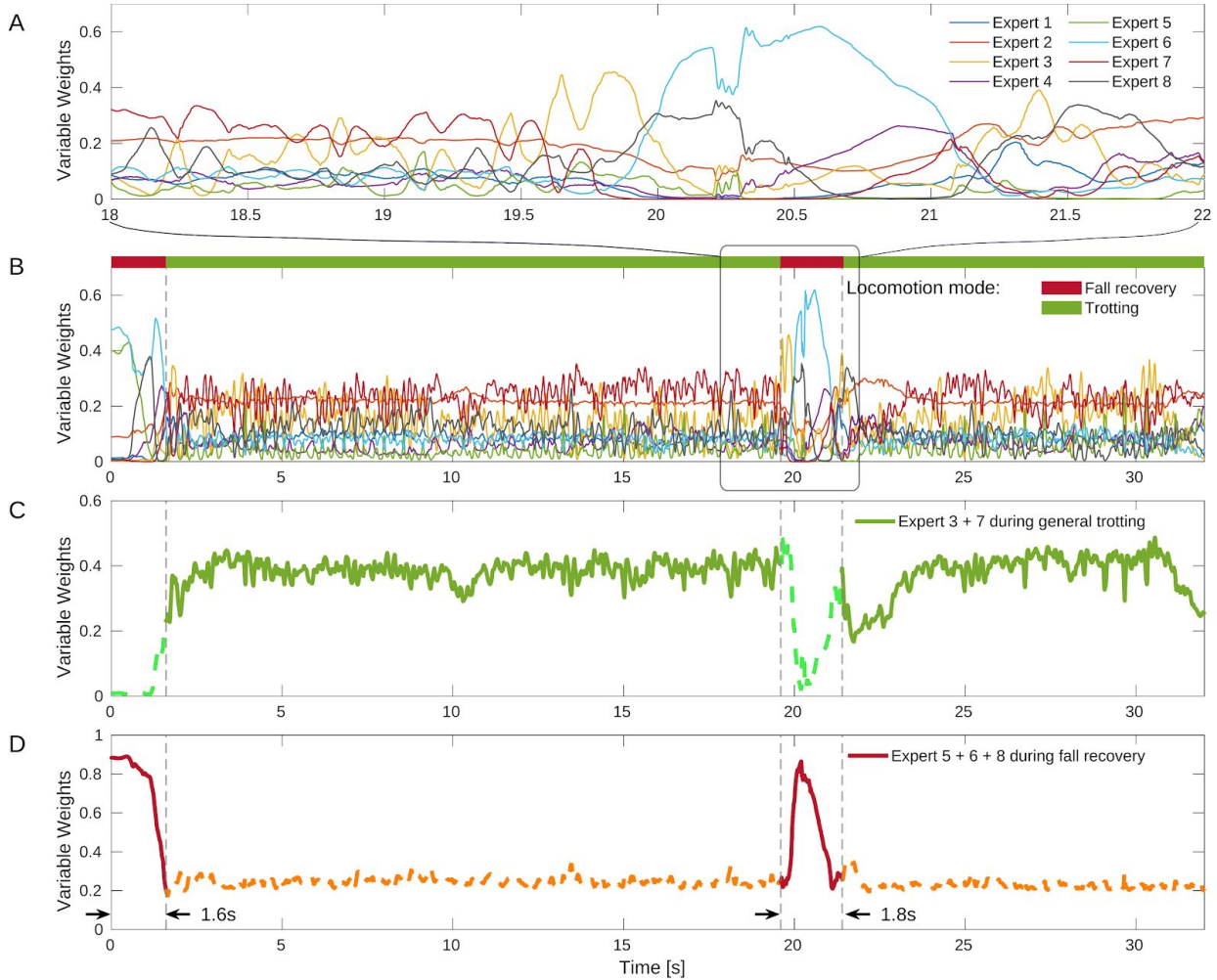

**Fig. S13. Continuous and variable weights of all experts during the real MELA experiment on grass (Fig. 5E).** (**A**) The variable activations within a zoomed period to show the transition of weights between multiple experts. (**B**) The variable activations of the entire multimodal locomotion with trotting and fall recovery during a target-following task. (**C-D**) The activation levels of paired weights from collaborating experts, where the expert groups (3, 7) and (5, 6, 8) cooperated together in trotting and fall recovery, respectively.

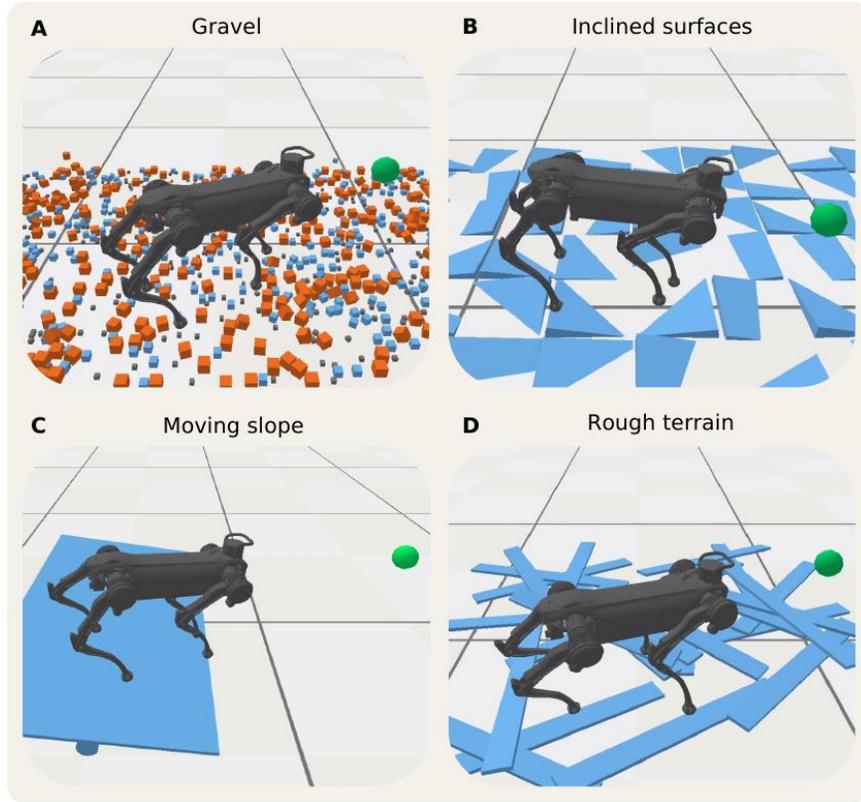

**Fig. S14. Four types of unseen terrains for testing the multi-skill MELA policy in the simulation.** (**A**) The gravel is constructed by a variety of freely moving **cubes** with dimensions of 0.02m, 0.035m, and 0.05m. (**B**) The inclined surfaces consist of rectangular slabs (0.4 m x 0.4 m x 0.2 m), which are statically placed with random orientations on the ground. (**C**) The moving slope has a changing inclination created by a seesaw with a maximum inclination of 0.17 *rad* (10 *deg*). (**D**) The rough terrain created by planks with the mass of 2.5kg and a size of 1.2 m x 0.12 m x 0.02 m randomly distributed on the ground.

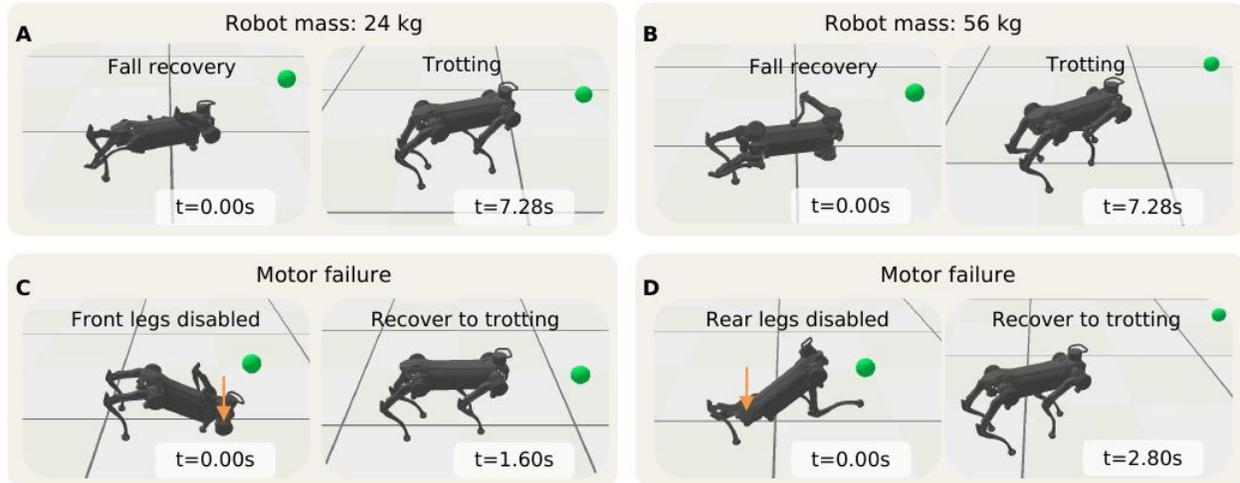

**Fig. S15. Simulated test scenarios for evaluating the robustness of the MELA policy.** (**A-B**) Uncertainties in dynamic properties are simulated by modifying the robot model, i.e., robot mass with variations of $\pm 25\%$, $\pm 30\%$ and $\pm 40\%$ of the original value (40kg). We show snapshots with the mass variation of $\pm 40\%$ as an extreme example, and the rest of the tests can be seen in movie S7. (**A**) Fall recovery and trotting with $60\%$ of the original mass. (**B**) Fall recovery and trotting with $140\%$ of the original mass. (**C-D**) Motor failures are emulated by disabling (zero torque) the front legs (**C**) and rear legs (**D**) respectively for one second. In both cases, the robot was able to recover from failures and accomplish the task.

**Fig. S16. Representative adaptive behaviour from the simulated scenario of steering on spot (fig. S4A, movie S6)**. (**A**) Snapshots depicting the behaviours during left and right steering. (**B**) Position references of all the joints during left and right steering phases. The smooth change in desired joint positions indicates that the MELA framework has learned how to synthesise expert skills during various transitions seamlessly.

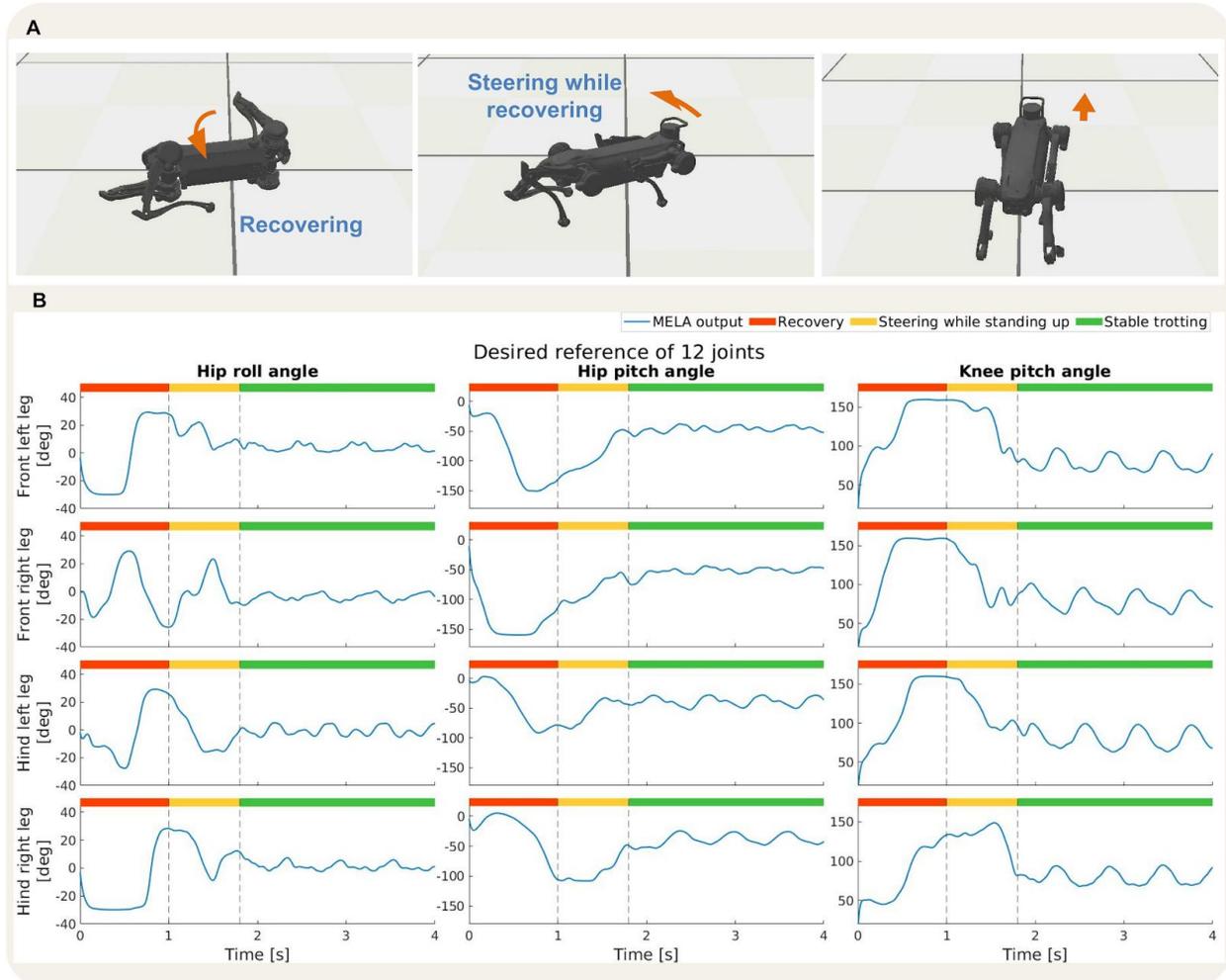

**Fig. S17. Representative adaptive behaviour from the simulated scenario of steering while recovering to trotting (fig. S4B, movie S6)**. (**A**) Snapshots depicting the behaviours during recovery and steering. (**B**) Position references of all the joints during recovery, steering, recovery, and trotting phases. The smooth change in desired joint positions indicates that the MELA framework has learned how to synthesise expert skills during various transitions seamlessly.

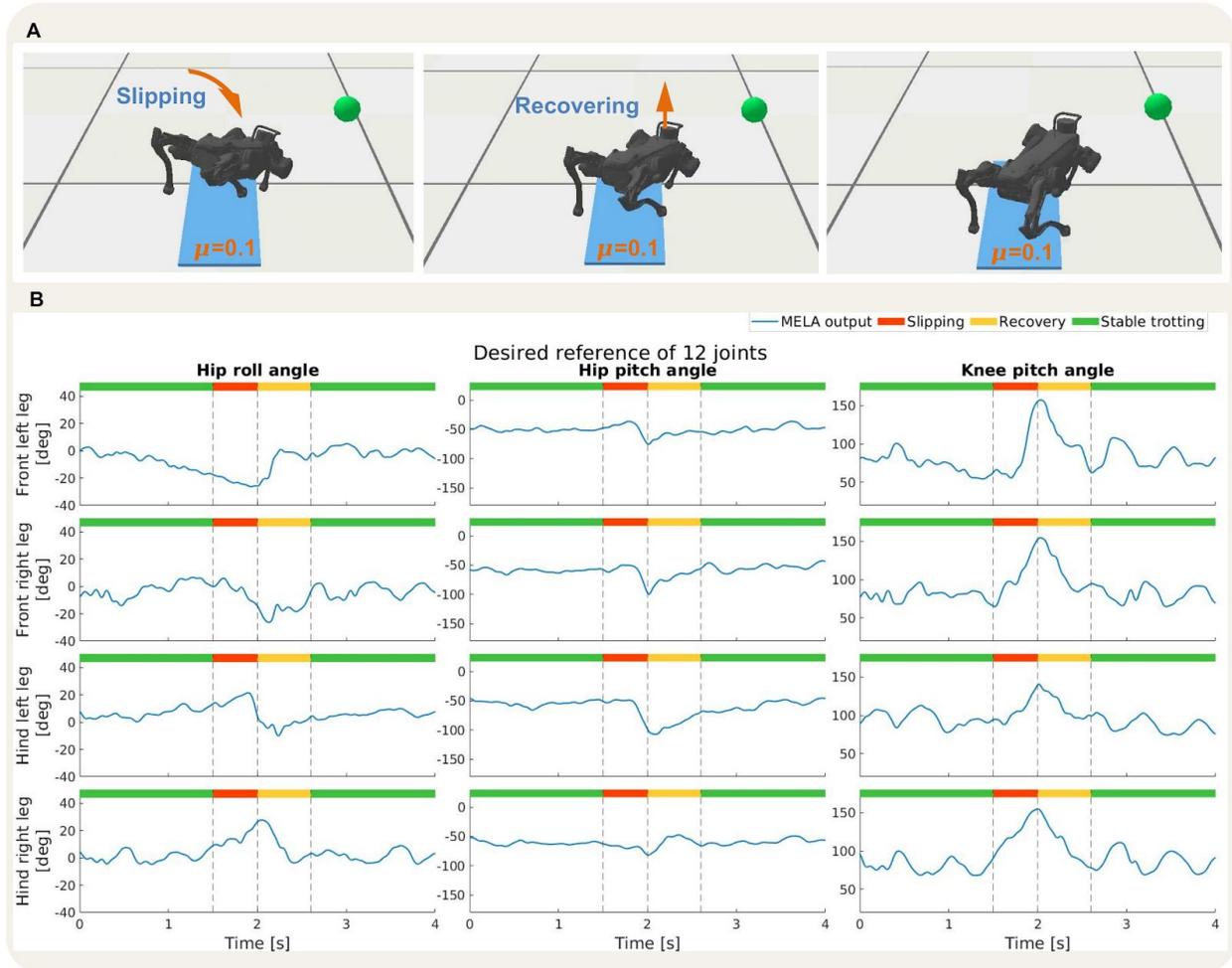

**Fig. S18. Representative adaptive behaviour from the simulated scenario of tripping (fig. S4C, movie S6)**. (**A**) Snapshots depicting the behaviours during slipping and recovery. (**B**) Position references of all joints during slipping, recovery, and trotting phases. The smooth change in desired joint positions indicates that the MELA framework has learned how to synthesise expert skills during various transitions seamlessly.

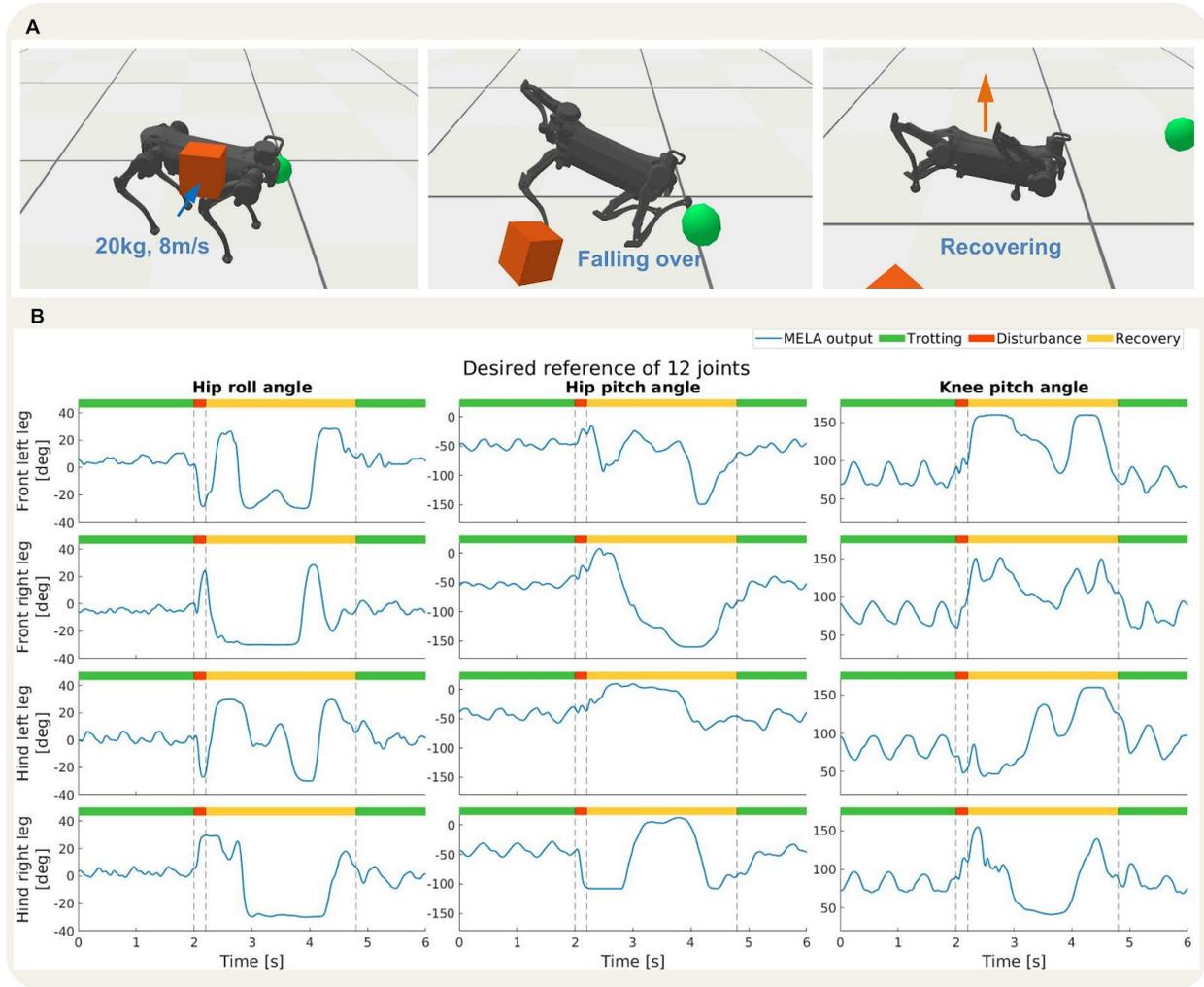

**Fig. S19. Representative adaptive behaviour from the simulated scenario of a large impact (fig. S4D, movie S6)**. (**A**) Snapshots depicting the behaviours during the moment of disturbance and recovery (**B**) Position references of all the joints during trotting, disturbance, and recovery phases. The smooth change in desired joint positions indicates that the MELA framework has learned how to synthesise expert skills during various transitions seamlessly.

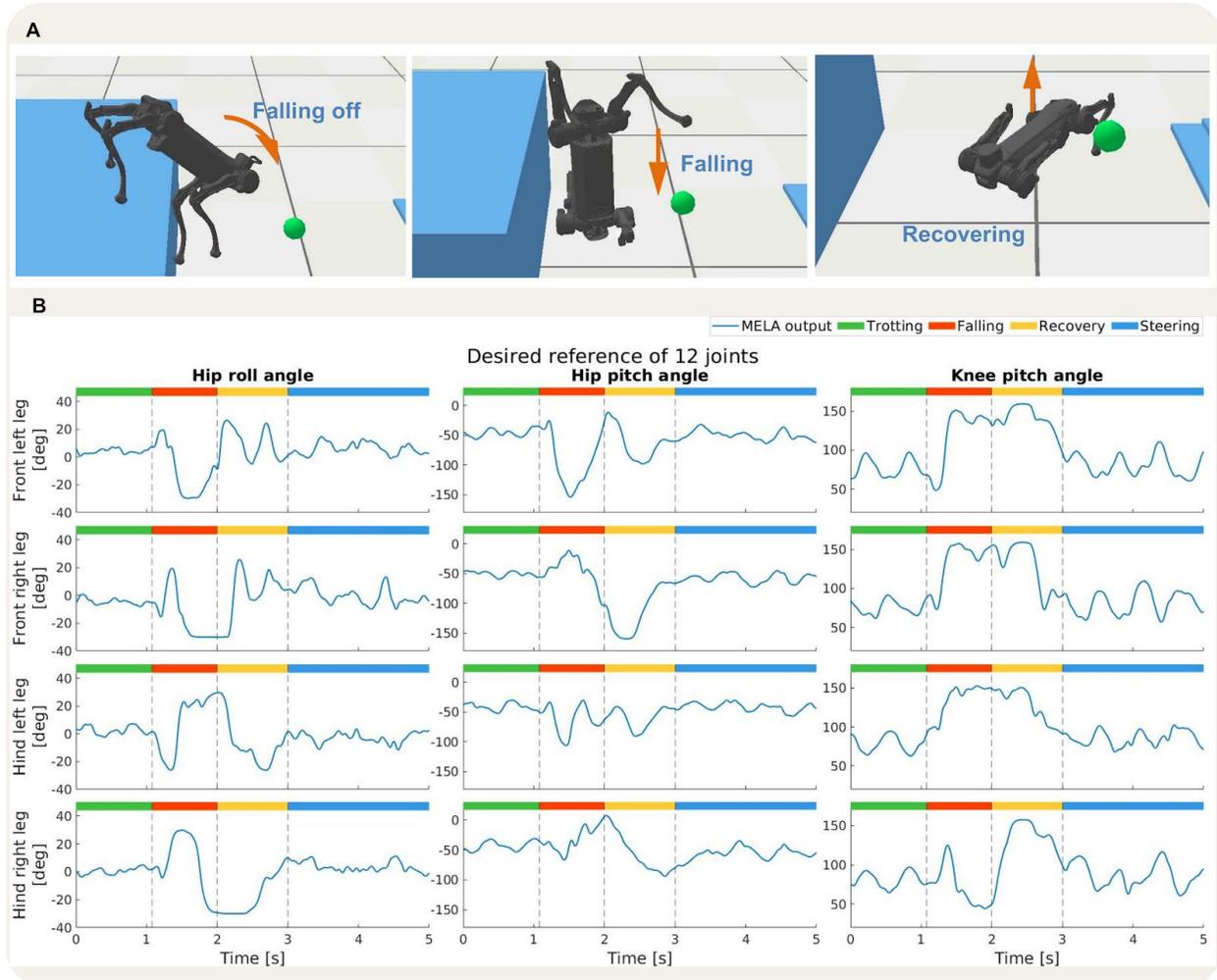

**Fig. S20. Representative adaptive behaviour from the simulated scenario of falling off a cliff (fig. S4E, movie S6)**. (**A**) Snapshots depicting the behaviours during falling and recovery. (**B**) Position references of all the joints during falling, recovery, and steering. The smooth change in desired joint positions indicates that the MELA framework has learned how to synthesise expert skills during various transitions seamlessly.

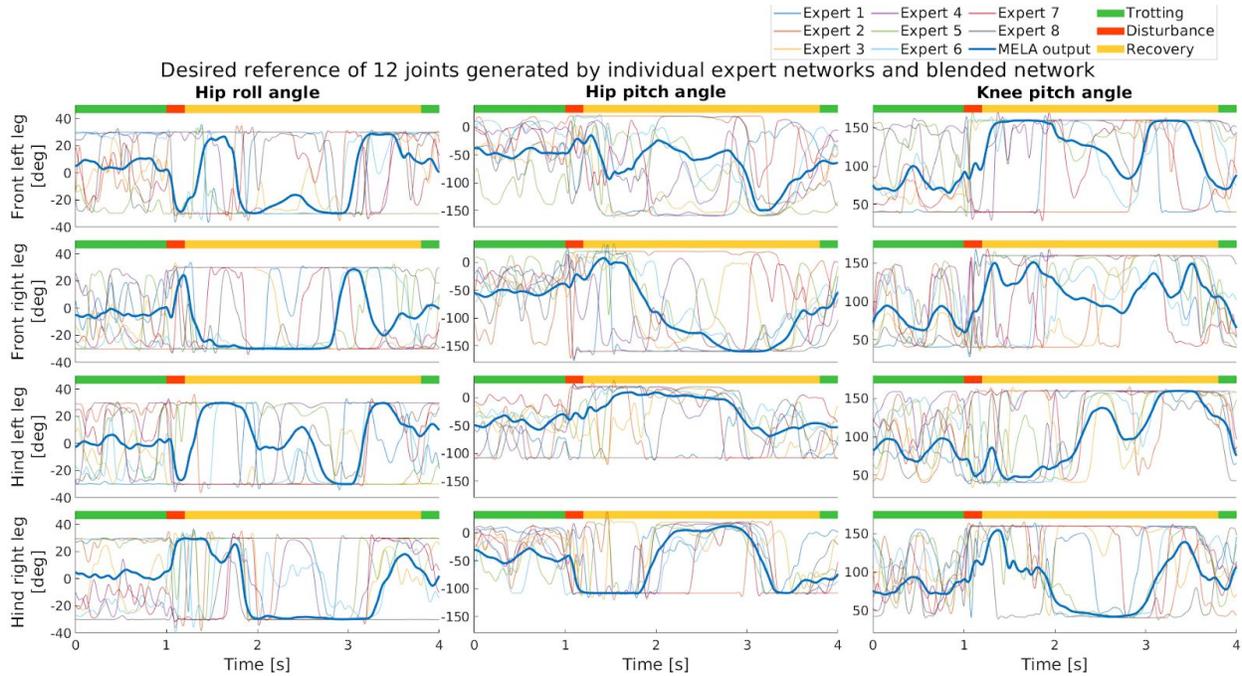

**Fig. S21. Analysis of responses from the MELA policy during the simulated scenario of a large external perturbation (fig. S4D, movie S6)**. We use the case of a flying box impact as a representative example to show how the policy actively reacts to perturbations with a smooth and seamless transition. The coloured bars at the top of each plot show the 3 phases during the cube impact scenario: the green, red, and yellow phases represent stable trotting, the time during the force impact by the high-speed cube, and the recovery process, respectively. In each subplot, the 8 semi-transparent lines are the outputs of each individual expert, and the blue solid line is the output of the synthesised MELA network. The outputs of the synthesised policy (solid blue lines) have very different characteristics from that of the 8 basic experts during all the phases, which suggests an interpolated behaviour and a nonlinear synthesis among the expert skills.

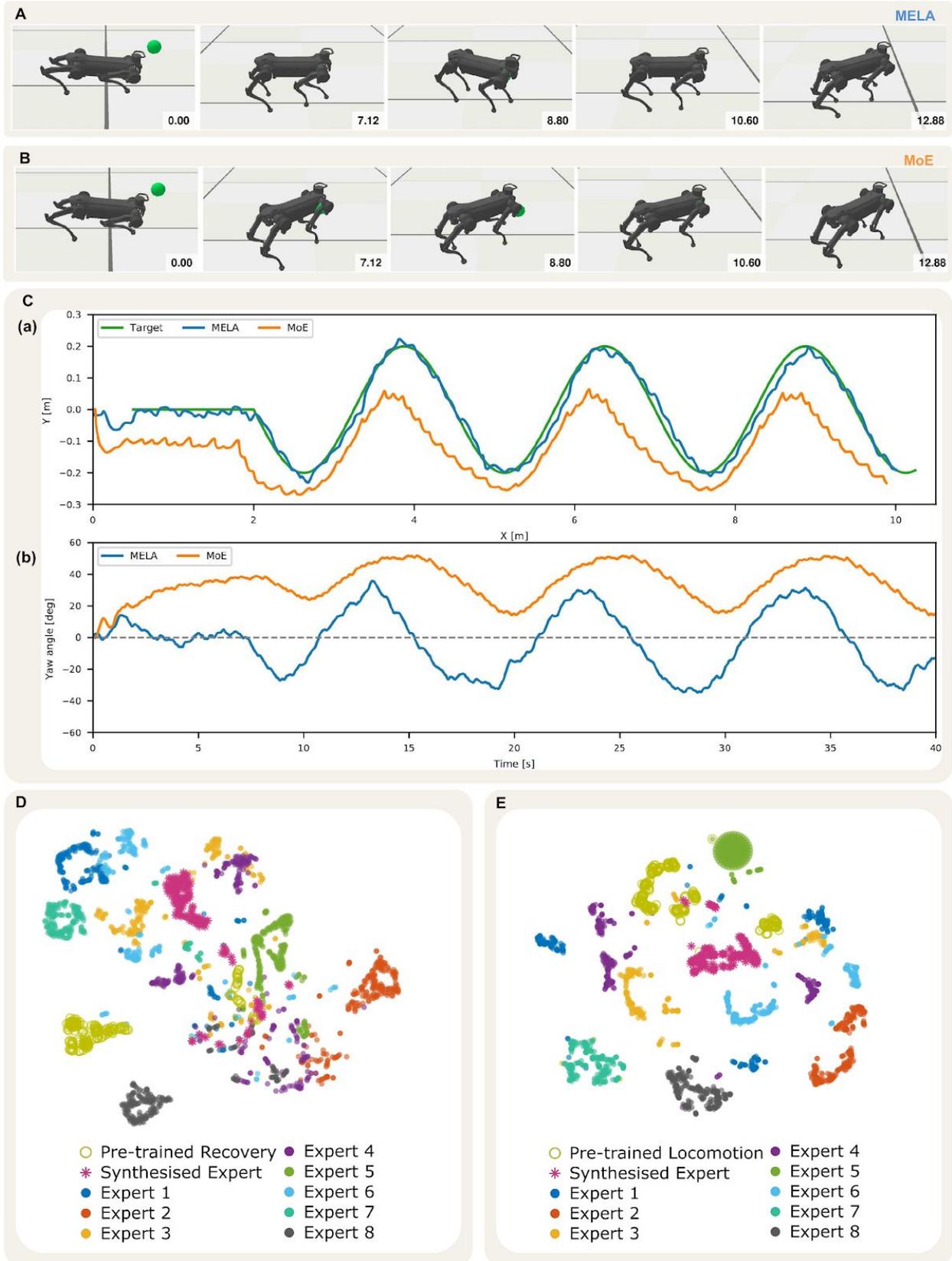

**Fig. S22. Phenomenon of asymmetric gait and imbalanced experts from MoE. (A)**

MELA policy performing a target-following test with fall recovery, steering and trotting. The robot was able to steer left and right to follow the moving target. (**B**) MoE policy performing a target-following test, where the yaw angle drifted due to the degenerated expert in steering skills. (**C**) Global position and yaw angle of the robot during the target-following test using MELA and MoE, respectively. (**D**) and (**E**) are the t-SNE analysis of MoE during fall recovery and trotting respectively, using target actions from pre-trained, co-trained, and synthesised experts. In contrast to the eight unique clusters from MELA (see Fig. 4D-E), the expert imbalance was observed in MoE that only half favoured experts were well trained with discernible clusters, while others are scattered and overlapped.

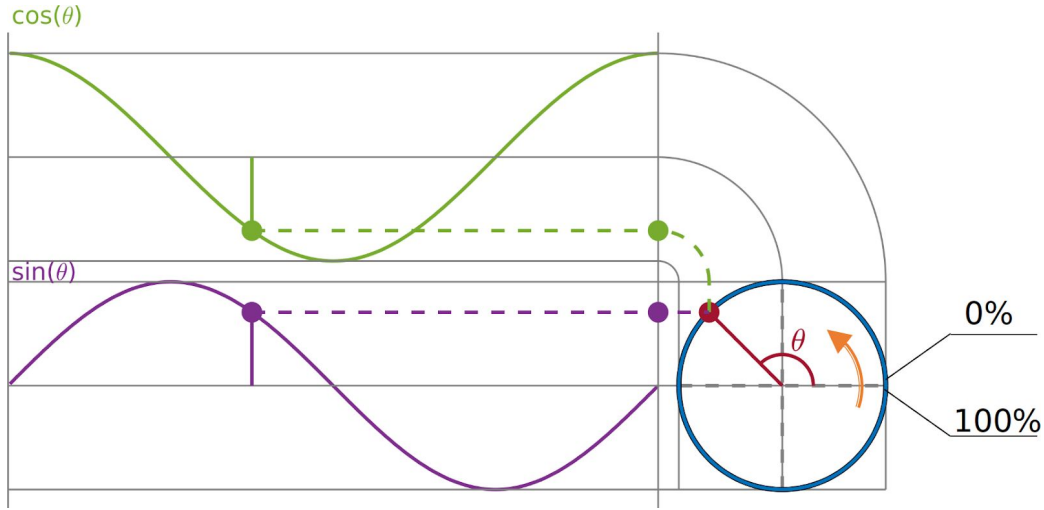

**Fig. S23. Illustration of the 2D phase vector for training the locomotion policy.** The sine and cosine functions are used to represent the time-varying phase variable in a continuous manner, and the resulting phase vector contains temporal information to describe the phase (0-100%) of a periodic gait.

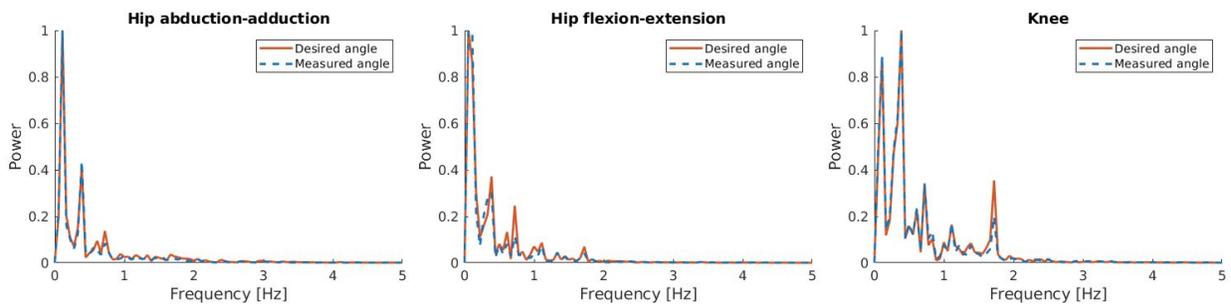

**Fig. S24. Normalised power spectrum analysis of motions during the real locomotion experiment (without the DC component).** The data were collected from the experiment shown in Fig. 5D. The majority of the frequency components were below 1Hz, and some small components were around 1.67Hz corresponding to the trotting motions, which indicated that all useful motion components were unaffected by the action filters.

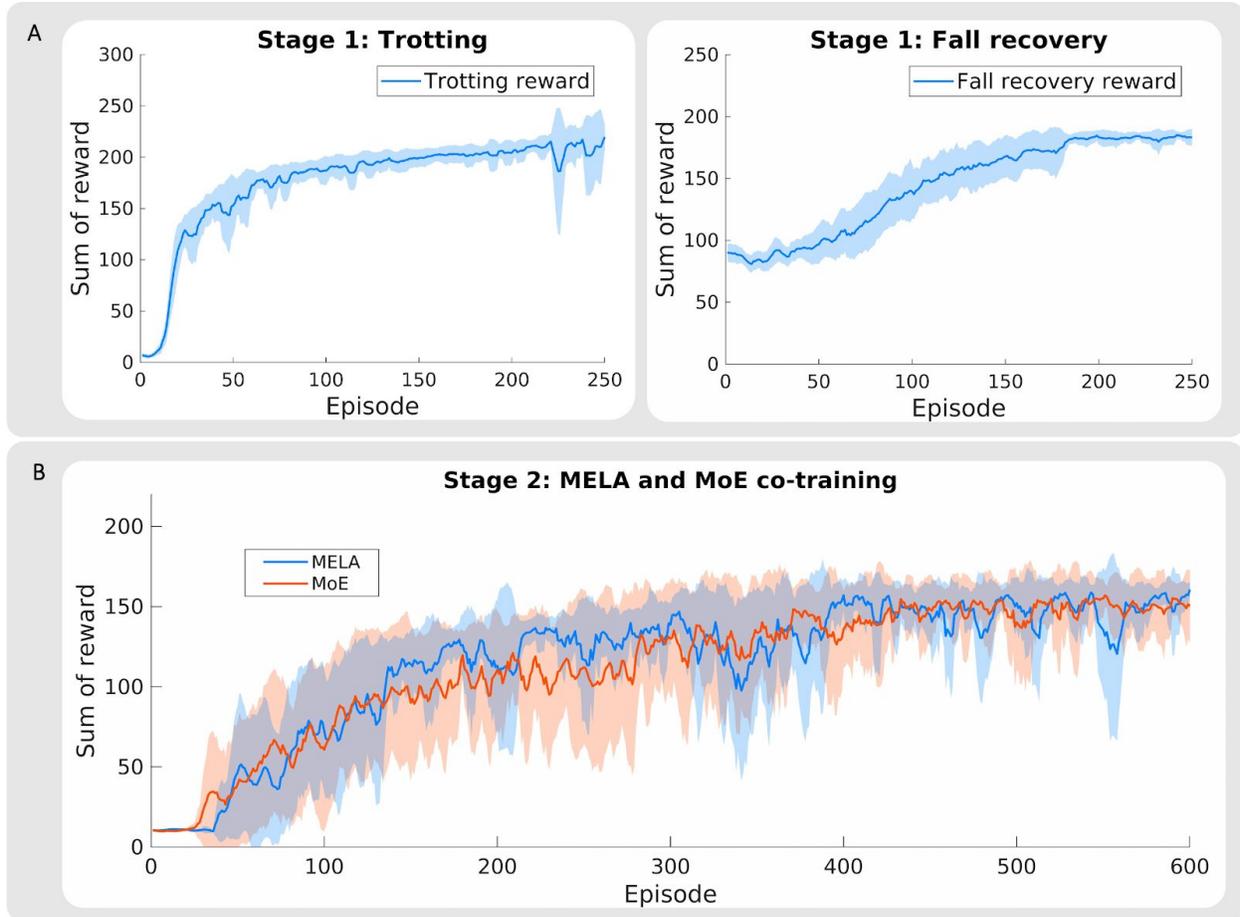

**Fig. S25. Learning curves during the 2-stage training of MELA and MoE.** (**A**) Learning curves of fall recovery and trotting experts. For training experts in the first stage, 250 episodes were required for both the fall recovery and trotting tasks. The same fall recovery and trotting experts can be used for the second stage of contraining for both MELA and MoE. (**B**) Learning curves of MELA and MoE during the second stage of co-training. The training continued until 600 episodes to ensure that each policy had fully converged to its maximum reward for a comparison study. Each episode consists of 5000 samples which were collected at 25 Hz.

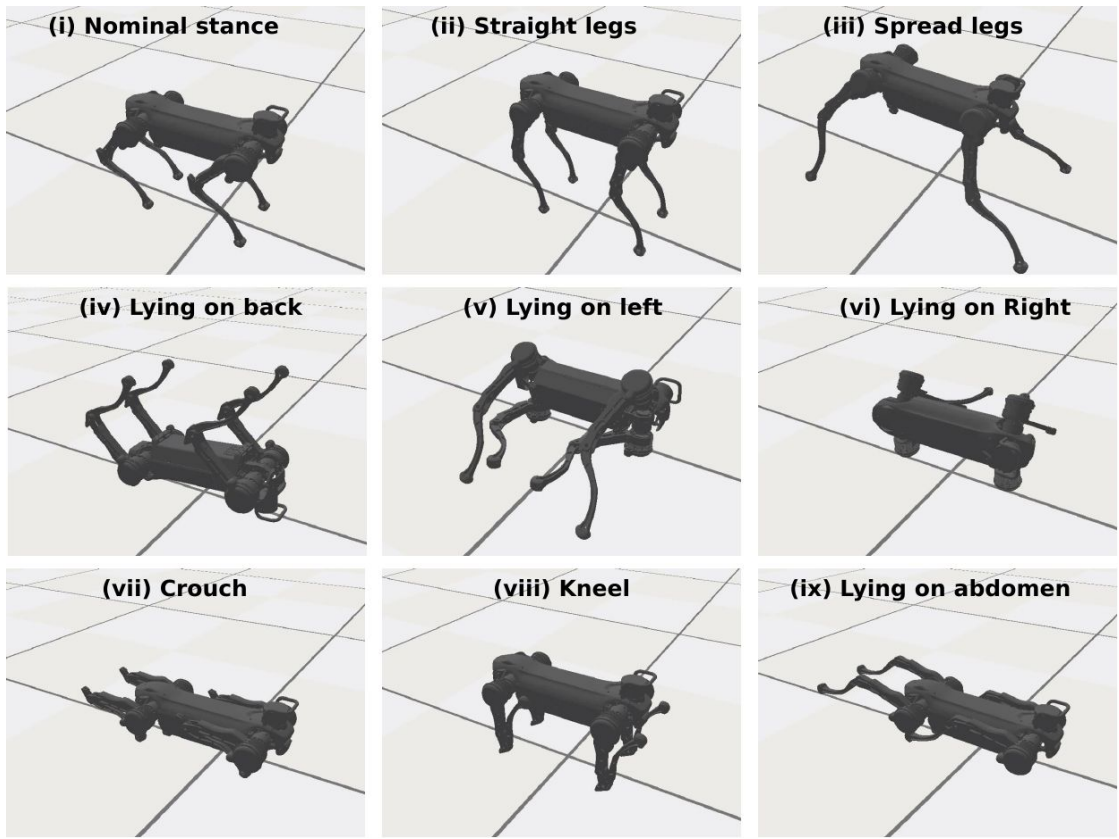

**Fig. S26. Nine distinct configurations used as the initialisation for training fall recovery policies in simulation**. Snapshots are taken from the physics-based simulator using the PyBullet engine (*64*).

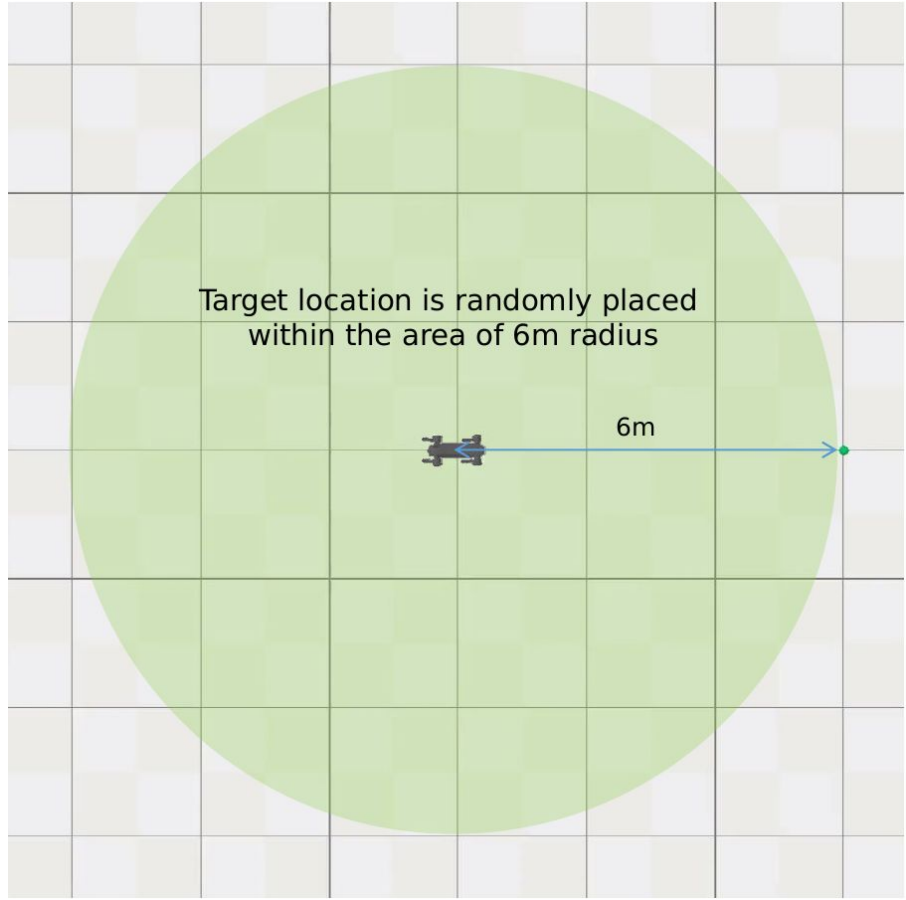

**Fig. S27. Setting of the target location for training MELA policies in simulation**. During the initialisation of each sample collection episode, the target location (the green ball) is randomly placed within the area of 6 m radius around the robot, and remains fixed within the same episode.

# Tables

**Table S1. Distribution matrix of expert specialisations over motor skills.** As shown in Fig. 4A, more than one expert is activated for a motor skill, and hence we highlight the top two dominant experts that are activated in each motor skill in this table.

| Motor Skill index | \ | Expert index | | | | | | | |
|---|---|---|---|---|---|---|---|---|---|
| | | 1 | 2 | 3 | 4 | 5 | 6 | 7 | 8 |
| I   | | |   |   |   | ● | ● |   |   |
| II  | | |   |   |   |   | ● |   | ● |
| III | | |   |   |   | ● | ● |   |   |
| IV  | | ● | ● |   |   |   |   |   |   |
| V   | | |   |   | ● |   |   | ● |   |
| VI  | | ● |   |   |   |   | ● |   |   |
| VII | | |   | ● |   |   |   | ● |   |
| VIII| | |   | ● |   |   |   | ● |   |

**Table S2. Specification of the *Jueying* quadruped robot.**

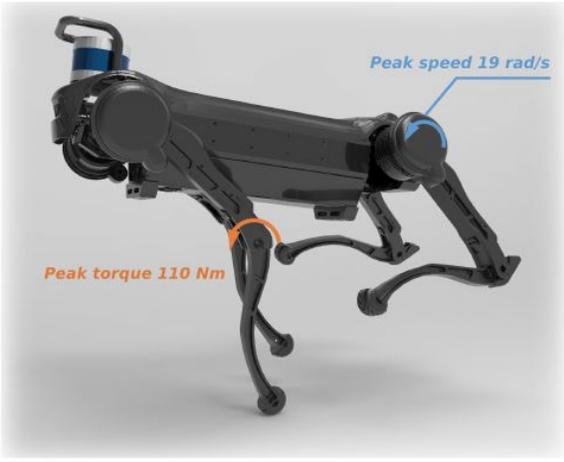

| | Joint range (°) | $\tau\ (Nm)$ [Peak] | $\omega\ (rad/s)$ [Peak] |
|---|---|---|---|
| Hip roll  | -25 – 25   | 25 [75]  | 13 [19] |
| Hip pitch | -159 – 29  | 25 [75]  | 13 [19] |
| Knee      | 37 – 164   | 45 [110] | 13 [19] |
| Body size (LWH) | | 0.85m×0.30m×0.30m | |
| Leg length (upper, lower) | | 0.33m, 0.34m | |

**Table S3. Detailed descriptions of the individual reward terms.**

| Physical quantities | Equations |
|---|---|
| Base pose | $\varphi(\phi, [0, 0, -1], \alpha)$, $\alpha = -2.35$ |
| Base height | $\varphi(h, \hat{h}, \alpha)$, $\alpha = -51.16$ |
| Base velocity | $\varphi(v_{base}^{world}, \hat{v}_{base}^{world}, \alpha)$, $\alpha = -18.42$ |
| Joint torque regularisation | $\varphi(\tau, 0, \alpha)$, $\alpha = -0.003$ |
| Joint velocity regularisation | $\varphi(\dot{q}, 0, \alpha)$, $\alpha = -0.026$ |
| Foot ground contact | $\begin{cases} 1, & \text{foot in contact with ground} \\ 0 & \end{cases}$ |
| Body ground contact | $\begin{cases} 0, & \text{main body in contact with ground} \\ 1 & \end{cases}$ |
| Yaw velocity | $\varphi(\omega, 0, \alpha)$, $\alpha = -7.47$ |
| Joint position reference | $\varphi(q, \hat{q}, \alpha)$, $\hat{q}$ is the joint reference, $\alpha = -29.88$ |
| Foot contact reference | $\begin{cases} 1, & \text{match desired foot contact in imitation data} \\ 0 & \end{cases}$ |
| Robot heading | $\varphi(u_{goal,base}^{base}, [1, 0, 0], \alpha)$, $\alpha = -2.35$ |
| Goal position | $\varphi(p_{goal}^{world}, p_{base}^{world}, \alpha)$, $\alpha = -0.74$ |
| Swing and stance | $\varphi\left( 1/4 \cdot \sum_{n=1}^{4} ((h_{foot,n}^{world} - \hat{h}_{foot,n}^{world}) \cdot v_{foot,n}^{world}), 0, \alpha \right)$, $\alpha = -460.50$ |
| Foot placement | $\varphi\left( 1/4 \cdot \sum_{n=1}^{4} p_{foot,n}^{world}, p_{base}^{world}, \alpha \right)$, $\alpha = -18.42$ |

**Table S4. Weights of the reward terms for different tasks.** Trotting and fall recovery used a subset of the reward terms, and the multimodal MELA locomotion used all the reward terms.

| Physical quantities | Trotting | Fall Recovery | MELA |
|---|---|---|---|
| Base pose | 0.071 | 0.333 | 0.100 |
| Base height | 0.036 | 0.333 | 0.100 |
| Base velocity | 0.178 | 0.067 | 0.071 |
| Joint torque regularisation | 0.018 | 0.067 | 0.020 |
| Joint velocity regularisation | 0.018 | 0.067 | 0.020 |
| Foot ground contact | 0.018 | 0.067 | 0.020 |
| Body ground contact | 0.018 | 0.067 | 0.020 |
| Yaw velocity | 0.071 | 0.00 | 0.020 |
| Foot clearance | 0.036 | 0.00 | 0.036 |
| Joint position reference | 0.416 | 0.00 | 0.167 |
| Foot contact reference | 0.083 | 0.00 | 0.033 |
| Average foot placement | 0.036 | 0.00 | 0.036 |
| Robot heading | 0.00 | 0.00 | 0.143 |
| Goal position | 0.00 | 0.00 | 0.214 |

**Table S5. Selection of state inputs for different tasks and neural networks.** The MELA gating network receives task-oriented state inputs (i.e., the normalised gravity vector, the angular and linear velocities, and the goal position) for the target-following locomotion; and the synthesised MELA network receives all the feedback except the goal position.

| Physical quantities | Trotting | Fall Recovery | MELA Gating Network | Synthesised MELA Network |
|---|---|---|---|---|
| **Joint position** | ✓ | ✓ | | ✓ |
| **Gravity vector** | ✓ | ✓ | ✓ | ✓ |
| **Angular velocity of the robot** | ✓ | ✓ | ✓ | ✓ |
| **Linear velocity of the robot in its local heading frame** | ✓ | | ✓ | ✓ |
| **Phase vector** | ✓ | | | ✓ |
| **Goal position** | | | ✓ | |

**Table S6. Proportional-Derivative parameters for the joint-level PD controller.**

| | Hip roll | Hip Pitch | Knee pitch |
|---|---|---|---|
| **Kp** (Nm/rad) | 700 | 700 | 700 |
| **Kd** (Nms/rad) | 10 | 10 | 10 |

**Table S7. Hyperparameters for SAC.**

| | |
|---|---|
| Smoothing loss coefficient | 2.0 |
| Learning rate | 3e-4 |
| Weight decay | 1e-6 |
| Discount factor | 0.987 |
| Soft target update | 0.001 |
| Replay buffer size | 1e6 |
| Steps per epoch | 5e3 |